\documentclass[10pt]{article} 
\usepackage[preprint]{tmlr}

\usepackage{amsmath,amsfonts,bm}

\def\eqref#1{equation~\ref{#1}}

\def\1{\bm{1}}

\DeclareMathAlphabet{\mathsfit}{\encodingdefault}{\sfdefault}{m}{sl}
\SetMathAlphabet{\mathsfit}{bold}{\encodingdefault}{\sfdefault}{bx}{n}

\PassOptionsToPackage{sort}{natbib}
\usepackage{amssymb,amsmath,amsthm}
\usepackage{graphicx}
\usepackage{microtype}
\usepackage{tabularx, colortbl, booktabs} %
\usepackage[table,xcdraw,dvipsnames]{xcolor}
\usepackage[normalem]{ulem}
\usepackage{subcaption}
\usepackage{times}
\usepackage{epsfig}
\usepackage{color}
\usepackage[T1]{fontenc}
\usepackage{textcomp}
\usepackage{array}
\usepackage{svg}
\usepackage{placeins}
\usepackage{import}
\usepackage{xifthen}
\usepackage{pdfpages}
\usepackage{transparent}
\usepackage{pifont}
\usepackage{xspace}
\usepackage{makecell}
\usepackage{etoolbox,xpatch}
\usepackage{rotating}
\usepackage{enumitem}
\usepackage{pdflscape}
\usepackage{titletoc}
\usepackage{wrapfig}
\usepackage{abbrv}
\usepackage{float}
\usepackage{algorithm}
\usepackage{algpseudocodex}
\usepackage{inconsolata}
\usepackage{tikz}
\usetikzlibrary{calc,tikzmark,arrows.meta} 
\tikzset{>={Stealth[length=2mm]}} 
\definecolor{GraphcoreCoral}{HTML}{FF6F79}
\colorlet{GraphcoreCite}{gray!60!GraphcoreCoral}
\colorlet{GraphcoreTakeawayFrame}{GraphcoreCoral}
\colorlet{GraphcoreTakeawayBack}{GraphcoreCoral!8!white}
\colorlet{GraphcoreTakeawayTitle}{GraphcoreCoral!90!black}
\colorlet{GraphcoreTakeawayTitleNormal}{GraphcoreTakeawayTitle!75!white}
\definecolor{GraphcoreWebBlue}{HTML}{B5E4EB}
\colorlet{GraphcoreFindingFrame}{GraphcoreWebBlue}
\colorlet{GraphcoreFindingBack}{GraphcoreWebBlue!17!white}
\definecolor{GraphcoreFindingTitle}{HTML}{1E808F}
\colorlet{GraphcoreFindingTitleNormal}{GraphcoreFindingTitle!75!white}
\colorlet{GraphcoreExampleFrame}{black!20!white}
\colorlet{GraphcoreExampleBack}{black!3!white}
\colorlet{GraphcoreExampleTitle}{black!65!white}
\definecolor{SpecDecDraftBox}{HTML}{F4FAF5}
\definecolor{SpecDecVerifyBox}{HTML}{F3F7FC}
\definecolor{SpecDecDraftLabel}{HTML}{3D8A55}
\definecolor{SpecDecVerifyLabel}{HTML}{3F73AD}
\definecolor{SpecDecRelaxedPi}{HTML}{9E1B1B}
\definecolor{TaxonomyTickGreen}{HTML}{23894F}
\definecolor{TaxonomyTickYellow}{HTML}{C48900}
\definecolor{SupergridAnnA}{HTML}{2563EB}
\definecolor{SupergridAnnB}{HTML}{0891B2}
\definecolor{SupergridAnnC}{HTML}{0D9488}
\definecolor{SupergridAnnD}{HTML}{16A34A}
\definecolor{SupergridAnnE}{HTML}{DB2777}
\definecolor{SupergridAnnF}{HTML}{6D28D9}
\newcommand{\taxcheck}{\textcolor{TaxonomyTickGreen}{\ding{51}}}
\newcommand{\taxsoftcheck}{\textcolor{TaxonomyTickYellow}{(\ding{51})}}
\newcommand{\taxhead}[1]{%
  \begingroup
  \renewcommand{\arraystretch}{0.88}%
  \makecell[l]{#1}%
  \endgroup
}
\newcommand{\taxnote}[1]{{\footnotesize\color{GraphcoreExampleTitle}$#1$}}
\usepackage[pagebackref,breaklinks=true,colorlinks]{hyperref}

\let\GraphcoreOrigSum\sum
\renewcommand{\sum}{\GraphcoreOrigSum\nolimits}
\hypersetup{
    colorlinks=true,
    citecolor=GraphcoreCite,
    breaklinks=true,
}
\usepackage{url}
\usepackage[all=normal, mathspacing=normal,mathdisplays=normal, floats=tight, paragraphs=tight, lists=normal]{savetrees}
\usepackage[capitalize]{cleveref}
\crefname{section}{Sec.}{Secs.}
\Crefname{section}{Section}{Sections}
\crefname{appendix}{Appendix}{Appendices}
\Crefname{appendix}{Appendix}{Appendices}
\crefname{table}{Tab.}{Tabs.}
\Crefname{table}{Table}{Tables}
\theoremstyle{plain}
\newtheorem{result}{Result}

\crefname{result}{Result}{Results}
\Crefname{result}{Result}{Results}
\crefname{lemma}{Lemma}{Lemmas}
\Crefname{lemma}{Lemma}{Lemmas}
\crefname{corollary}{Cor.}{Cors.}
\Crefname{corollary}{Corollary}{Corollaries}
\crefname{claim}{Claim}{Claims}
\Crefname{claim}{Claim}{Claims}

\usepackage[most]{tcolorbox}

\definecolor{DarkTurquoise}{RGB}{0,128,128}   

\newcounter{takeaway}
\newcounter{finding}
\crefname{takeaway}{Takeaway}{Takeaways}
\Crefname{takeaway}{Takeaway}{Takeaways}
\crefname{finding}{Finding}{Findings}
\Crefname{finding}{Finding}{Findings}

\tcbset{
  editorialbox/.style={
    enhanced,
    breakable,
    sharp corners,
    boxrule=0pt,
    frame hidden,
    fontupper=,
    fonttitle=\bfseries,
    attach title to upper={\ },
    before skip=3pt plus 1pt minus 1pt,
    after skip=3pt plus 1pt minus 1pt,
    left=5pt,
    right=3pt,
    top=3pt,
    bottom=3pt,
    boxsep=1pt,
    borderline east={0.25pt}{0pt}{black!8!white},
  },
  takeawaystyle/.style={
    editorialbox,
    colback=GraphcoreTakeawayBack,
    coltitle=GraphcoreTakeawayTitleNormal,
    borderline west={2pt}{0pt}{GraphcoreTakeawayFrame},
  },
  elevatedtakeawaystyle/.style={
    takeawaystyle,
    coltitle=GraphcoreTakeawayTitle,
    fonttitle=\bfseries,
    borderline west={2.8pt}{0pt}{GraphcoreTakeawayFrame},
  },
  findingstyle/.style={
    editorialbox,
    colback=GraphcoreFindingBack,
    coltitle=GraphcoreFindingTitleNormal,
    borderline west={2pt}{0pt}{GraphcoreFindingFrame},
  },
  elevatedfindingstyle/.style={
    findingstyle,
    coltitle=GraphcoreFindingTitle,
    fonttitle=\bfseries,
    borderline west={2.8pt}{0pt}{GraphcoreFindingFrame},
  },
}
\newenvironment{takeaway}[2][]{%
  \refstepcounter{takeaway}\label{#2}%
  \ifstrequal{#1}{elevated}{%
    \begin{tcolorbox}[elevatedtakeawaystyle,title={Key Takeaway~\thetakeaway:}]%
  }{%
    \begin{tcolorbox}[takeawaystyle,title={Takeaway~\thetakeaway:}]%
  }%
}{\end{tcolorbox}}
\newenvironment{finding}[2][]{%
  \refstepcounter{finding}\label{#2}%
  \ifstrequal{#1}{elevated}{%
    \begin{tcolorbox}[elevatedfindingstyle,title={Key Finding~\thefinding:}]%
  }{%
    \begin{tcolorbox}[findingstyle,title={Finding~\thefinding:}]%
  }%
}{\end{tcolorbox}}
\newtcolorbox{examplebox}{
  enhanced,
  colback=GraphcoreExampleBack,
  colframe=GraphcoreExampleFrame,
  fonttitle=\bfseries,
  coltitle=GraphcoreExampleTitle,
  fontupper=\small,
  title={Rendered prompt contents:},
  attach title to upper={\par\smallskip},
  boxrule=0.5pt,
  arc=3pt,
  left=3pt, right=3pt, top=2pt, bottom=2pt,
}
\newcommand{\promptquestion}[1]{%
  \par\smallskip
  \noindent{\setlength{\fboxsep}{4pt}\colorbox{black!12!white}{%
  \parbox{\dimexpr\linewidth-2\fboxsep\relax}{\textbf{Question.} #1}}}%
  \par\smallskip
}

\AtBeginDocument{%
  \setlength{\abovedisplayskip}{6pt plus 2pt minus 3pt}%
  \setlength{\belowdisplayskip}{6pt plus 2pt minus 3pt}%
  \setlength{\abovedisplayshortskip}{2pt plus 2pt minus 2pt}%
  \setlength{\belowdisplayshortskip}{3pt plus 2pt minus 2pt}%
}
\makeatletter
\renewcommand{\paragraph}{%
  \@startsection{paragraph}{4}{\z@}%
    {0pt}
    {-1em}
    {\normalsize\bfseries}%
}
\makeatother
\title{A Practical Investigation of Training-free\\Relaxed Speculative Decoding}

\author{\name Guoxuan Xia, Luka Ribar, \email g.xia21@imperial.ac.uk \\
\name  Paul Balanca \email paulb@graphcore.ai \\
      \addr Work done at Graphcore}

\def\month{MM}  
\def\year{YYYY} 
\def\openreview{\url{https://openreview.net/forum?id=XXXX}} 

\makeatletter

\let\oldmaketitle\maketitle
\renewcommand{\maketitle}{%
  \vspace*{-8mm}%
  \oldmaketitle
}
\makeatother

\begin{document}

\maketitle

\begin{abstract}

Speculative decoding accelerates sampling from an autoregressive LLM by using a faster auxiliary model to draft tokens which are then verified in parallel by the LLM. Standard speculative decoding is lossless: its rejection and resampling steps exactly preserve the LLM's sampling distribution. Recent work argues that \textit{relaxing} this strict guarantee can yield further speed-ups, controlled capability-speed \textit{trade-offs}, or even \textit{capability gains}. We practically investigate training-free relaxed speculative decoding techniques, unify existing approaches within a shared framework, benchmark them on contemporary settings, and distil takeaways and empirical findings for practitioners. Important takeaways include: relaxation can require \textit{considerable capability evaluation} unlike lossless speculative decoding, and many relaxed approaches rely on a drafter that is a good language model, making them unsuited for lightweight dedicated multi-token-prediction drafters. 
Code is available at: \url{https://github.com/graphcore-research/relaxed-spec-dec-repro}.
\end{abstract}
\vspace{-5mm}
\section{Introduction}

Large language models (LLMs) dominate modern applications of machine learning \citep{sajadieh2026aiindex}. Language modelling as a task is itself in turn dominated by \textit{autoregressive} (\texttt{AR}) models \citep{transformer,gpt3}, \ie models that predict one token after another \textit{in series} given previous tokens. Such \texttt{AR} models represent the frontier of capability, enabling transformative outcomes in programming, scientific research, and many other applications \citep{sajadieh2026aiindex}. A core issue, however, is the inherent slowness of generation that arises from the requirement to \textit{sequentially} produce tokens, especially in an era of abundant \textit{parallel} compute resources \citep{you2026globalaicomputingcapacity}. Applications often require lower latencies in order to satisfy user requirements \citep{anthropic_haiku45_2025,openai_codex_spark_2026} -- intuitively, snappier LLM responses enable greater human productivity.

One solution to this problem is speculative decoding (\texttt{spec-dec}) \citep{spec-dec}, which is motivated by the idea that for text generation, weaker but faster models may still be able to ``speculate'' decent short-range spans of text. Given an \texttt{AR} LLM, \texttt{spec-dec} employs a low-latency auxiliary model to ``draft'' tokens quickly, which are then ``verified'' \textit{in parallel} using the LLM. This better utilises parallel compute and reduces overall latency. A key feature of this algorithm is that ``bad'' draft tokens are stochastically rejected and resampled during verification in such a way that \textit{strictly preserves} the LLM's output distribution, giving theoretically \textit{lossless} acceleration. Speculative decoding is seeing increasing adoption across a wide range of models, such as DeepSeek and Qwen \citep{liu2024deepseekv3technicalreport,qwen35}, and inference frameworks such as vLLM and SGLang \citep{kwon2023efficient,zheng2024sglang,schmittulms2025speculators}.

Recently, a new body of research has argued that the strict distribution preservation of speculative decoding is disadvantageous; generally, it proposes to \textit{relax} this requirement in favour of sampling from an alternative distribution to the original LLM. There are a number of motivations for doing this, for example, to enable the option of \textit{trading off} task capability for speed-up \citep{holsman-etal-2025-fuzzy}, increasing speed-up by accepting semantically valid tokens that strict speculative decoding would reject \citep{bachmann2025judge}, or even improving task capability \citep{yuan-etal-2024-spec-cont-dec}. In this work, we perform a practical investigation of such relaxed speculative decoding approaches (in particular, those that are training-free). We aim to better inform practitioners and researchers about the \textit{practical utility} of such approaches, given the variety of methods and experimental settings in recent literature. To this end, we present the following \textbf{key contributions}:
\begin{enumerate}[left=0pt, topsep=0pt,itemsep=-1ex,partopsep=1ex,parsep=1ex]
    \item We provide a primer on strict speculative decoding (\texttt{strict} \texttt{spec-dec}), presenting the algorithm and explaining \textit{why} and \textit{when} it produces latency speed-ups. We also discuss how speed-up is measured and modelled, as well as how inference infrastructure (\ie the hardware-software stack) can affect realisable speed-up.
    \item We build a \textit{taxonomy} over the literature of relaxed speculative decoding (\texttt{relaxed} \texttt{spec-dec}), organising diverse methods and narratives under a unified framework. The aim is for the reader to better understand not only how existing approaches are concretely implemented, but also their motivating scenarios, and how they relate to each other.
    \item We benchmark a range of relaxed speculative decoding approaches on modern drafter-verifier pairs and reasoning benchmarks over different inference-time parameter settings. This allows us to not only fairly compare approaches amongst themselves, but also against their respective paper narratives. This is important as the literature suffers generally from narrow and disparate experimental settings that preclude easy comparison.
    \item In relation to the problem setting of (relaxed) \texttt{spec-dec}, we extract practical \textcolor{GraphcoreTakeawayTitle}{\textbf{Takeaway}}s for the reader. Based on our experimental investigation of \texttt{relaxed} \texttt{spec-dec} methods, we also present practical empirical \textcolor{GraphcoreFindingTitle}{\textbf{Finding}}s.  
\end{enumerate}

\section{Preliminaries}
\begin{figure}[t]
\vspace{-10mm}
\centering
\begin{minipage}[t]{0.49\linewidth}
\begin{algorithm}[H]
\caption{\tt{strict spec-dec}}\label{alg:strict-spec-dec}
\small
\begin{algorithmic}
\Require draft model $q$, verifier model $p$, \\context $x_{<t}$, draft length $N_{\text{draft}}$
\BeginBox[fill=SpecDecDraftBox]
\LComment{\textcolor{SpecDecDraftLabel}{\textbf{DRAFT}}}
\LComment{Sample $N_{\text{draft}}$ tokens (auto-regressively) from $q$}
\For{$i = 0$ to $N_{\text{draft}}-1$}
    \State $q_{t+i}(\cdot) = q(\cdot|x_{<t+i})$
    \State $x_{t+i} \sim q_{t+i}(\cdot)$
\EndFor
\EndBox
\BeginBox[fill=SpecDecVerifyBox]
\LComment{\textcolor{SpecDecVerifyLabel}{\textbf{VERIFY}}}
\LComment{Run $p$ in parallel to score the draft tokens}
\State $p_{t+i}(\cdot) = p(\cdot|x_{<t+i})$ \quad for $i \in \{0,\ldots,N_{\text{draft}}\}$
\LComment{Accept or reject draft tokens}
\State $a_i \sim \operatorname{Ber}\!\left(\min\!\left\{1,
    \frac{p_{t+i}(x_{t+i})}{q_{t+i}(x_{t+i})}
\right\}\right)$
\Statex \hspace{\algorithmicindent}$\forall i \in \{0,\ldots,N_{\text{draft}}-1\};\quad a_{N_{\text{draft}}}=0$
\LComment{Find the earliest draft rejection position}
\State $i_\text{end} = \min\{i \in \{0,\ldots,N_{\text{draft}}\} : a_i = 0\}$
\If{$i_\text{end} < N_{\text{draft}}$}
\LComment{If draft rejection occurs}
    \LComment{Sample a new token from the residual distribution}
    \State $p^{\text{res}}(\cdot) =\operatorname{norm}\!\left(\max\{0,p_{t+i_\text{end}}(\cdot)-q_{t+i_\text{end}}(\cdot)\}\right)$
    \State Sample $x_{t+i_\text{end}} \sim p^{\text{res}}(\cdot)$
\Else
    \LComment{All draft tokens accepted}
    \LComment{Sample bonus next token after end of draft from $p$}
    \State $p^{\text{bonus}}(\cdot) = p_{t+N_{\text{draft}}}(\cdot)$
    \State Sample $x_{t+i_\text{end}} \sim p^{\text{bonus}}(\cdot)$
\EndIf
\EndBox
\LComment{Return accepted draft tokens + additional token}
\State \Return $x_t,\ldots,x_{t+i_\text{end}}$
\end{algorithmic}
\end{algorithm}
\end{minipage}
\hfill
\begin{minipage}[t]{0.49\linewidth}
\begin{algorithm}[H]
\caption{\tt{relaxed spec-dec}}\label{alg:relaxed-spec-dec}
\small
\begin{algorithmic}
\Require $q$, $p$, $x_{<t}$, $N_{\text{draft}}$, relaxation parameter $\textcolor{SpecDecRelaxedPi}{\alpha}$,\\ relaxed target distributions $\textcolor{SpecDecRelaxedPi}{\pi^{\text{rej}}}, \textcolor{SpecDecRelaxedPi}{\pi^{\text{res}}}, \textcolor{SpecDecRelaxedPi}{\pi^{\text{bonus}}}$
\BeginBox[fill=SpecDecDraftBox]
\LComment{\textcolor{SpecDecDraftLabel}{\textbf{DRAFT}}}
\For{$i = 0$ to $N_{\text{draft}}-1$}
    \State $q_{t+i}(\cdot) = q(\cdot|x_{<t+i})$
    \State $x_{t+i} \sim q_{t+i}(\cdot)$
\EndFor
\EndBox
\BeginBox[fill=SpecDecVerifyBox]
\LComment{\textcolor{SpecDecVerifyLabel}{\textbf{VERIFY}}}
\State $p_{t+i}(\cdot) = p(\cdot|x_{<t+i})$ \quad for $i \in \{0,\ldots,N_{\text{draft}}\}$
\LComment{Construct relaxed target distributions $\textcolor{SpecDecRelaxedPi}{\pi}$}
\State $\textcolor{SpecDecRelaxedPi}{\pi^{\text{rej}}_{t+i}(\cdot)} = \textcolor{SpecDecRelaxedPi}{\pi^{\text{rej}}[q_{t+i},p_{t+i},x_{t+i},\alpha](\cdot)}$
\State $\textcolor{SpecDecRelaxedPi}{\pi^{\text{res}}_{t+i}(\cdot)} = \textcolor{SpecDecRelaxedPi}{\pi^{\text{res}}[q_{t+i},p_{t+i},x_{t+i}, \alpha](\cdot)}$
\Statex \hspace{\algorithmicindent}$\text{for } i \in \{0,\ldots,N_{\text{draft}}\}$
\LComment{Replace rejection, residual and bonus distributions with $\textcolor{SpecDecRelaxedPi}{\pi}$}
\State $a_i \sim \operatorname{Ber}\!\left(\min\!\left\{1,
    \frac{\textcolor{SpecDecRelaxedPi}{\pi^{\text{rej}}_{t+i}(x_{t+i})}}{q_{t+i}(x_{t+i})}
\right\}\right)$
\Statex \hspace{\algorithmicindent}$\forall i \in \{0,\ldots,N_{\text{draft}}-1\};\quad a_{N_{\text{draft}}}=0$
\State $i_\text{end} = \min\{i \in \{0,\ldots,N_{\text{draft}}\} : a_i = 0\}$
\If{$i_\text{end} < N_{\text{draft}}$}
    \State $p^{\text{res}}(\cdot) =\operatorname{norm}\!\left(\max\{0,\textcolor{SpecDecRelaxedPi}{\pi^{\text{res}}_{t+i_\text{end}}(\cdot)}-q_{t+i_\text{end}}(\cdot)\}\right)$
    \State Sample $x_{t+i_\text{end}} \sim p^{\text{res}}(\cdot)$
\Else
    \LComment{Optional drafter step if needed for $\textcolor{SpecDecRelaxedPi}{\pi^{\text{bonus}}}$}
    \State $q_{t+N_{\text{draft}}}(\cdot) = q(\cdot|x_{<t+N_{\text{draft}}})$ 
    \State $\textcolor{SpecDecRelaxedPi}{\pi^{\text{bonus}}_{t+N_{\text{draft}}}(\cdot)} = \textcolor{SpecDecRelaxedPi}{\pi^{\text{bonus}}[q_{t+N_{\text{draft}}},p_{t+N_{\text{draft}}},x_{t+N_{\text{draft}}},\alpha](\cdot)}$
    \State $p^{\text{bonus}}(\cdot) = \textcolor{SpecDecRelaxedPi}{\pi^{\text{bonus}}_{t+N_{\text{draft}}}(\cdot)}$
    \State Sample $x_{t+i_\text{end}} \sim p^{\text{bonus}}(\cdot)$
\EndIf
\EndBox
\State \Return $x_t,\ldots,x_{t+i_\text{end}}$
\end{algorithmic}
\end{algorithm}
\end{minipage}
\caption{\textbf{Left}: In strict speculative decoding tokens are rapidly drafted and then verified in parallel. Draft tokens are stochastically rejected; the first rejected token is resampled such that overall sampling is from verifier target distribution $p$. \textbf{Right}: Relaxed speculative decoding can replace $p$ with relaxed target distribution $\pi$ at different points.}
\vspace{-3mm}
\end{figure}
We provide a technical introduction to strict speculative decoding and then introduce a unified framework for relaxed speculative decoding. A \textbf{glossary of notation} can be found in \cref{app:gloss} for the benefit of the reader.

\subsection{Speculative Decoding (\texttt{strict} \texttt{spec-dec})}\label{sec:spec-dec}
As previously mentioned, many applications of machine learning are dominated by autoregressive (\texttt{AR}) LLMs that necessarily need to generate one token after the other in sequence, and this is an issue as lower generation latency is important for many applications. At the same time, \texttt{AR} LLM decoding is generally \textit{memory bound}, that is to say, latency is dominated by repeatedly moving model weights and KV-cache activations, leaving available parallel compute underutilised \citep{ma2026infhard}. The consequence of this spare compute capacity is that increasing parallelism over tokens (\eg via batching) has a sublinear, often very small, impact on latency. 

Following from the above, speculative decoding \citep{spec-dec} is an algorithm that aims to better utilise available compute resources via parallelisation in order to reduce generation latency. Importantly, it provably samples from the same distribution as the LLM-to-be-accelerated, \ie it is \textit{lossless} acceleration. \cref{alg:strict-spec-dec} describes the approach: a fast auxiliary drafter model $q$ rapidly proposes an $N_\text{draft}$-long segment, which is then verified in parallel by the LLM-to-be-accelerated $p$ (the verifier). Each draft token $x\in \{0,\dots,V-1\}$, where $V$ is the vocab size, is stochastically accepted/rejected based on Bernoulli random variable $a \sim \operatorname{Ber}\!\left(\min\!\left\{1,p(x)/q(x)\right\}\right)$. At the first rejection ($a=0$), the rejected token is resampled from the residual distribution $\operatorname{norm}\!\left(\max\{0,p(\cdot)-q(\cdot)\}\right)$, where $\operatorname{norm}$ simply normalises to a valid probability distribution, recovering the distribution of $p$ losslessly. A detail that is sometimes overlooked is that the verifier $p$ also produces next-token logits for the final drafter token. If all draft tokens are accepted, then a bonus token is sampled from these logits. In fact, amongst algorithms that sample from $p$ by drafting a token from $q$, and then randomly rejecting and resampling, the above achieves the lowest possible expected rejection rate \citep{spec-dec-theory}.

\begin{figure}[t]
\vspace{-10mm}
    \centering
\includegraphics[width=.49\linewidth]{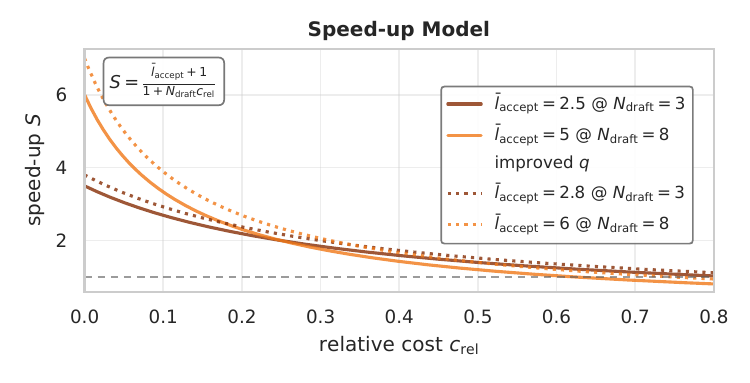}
        \hfill
    \includegraphics[width=.49\linewidth]{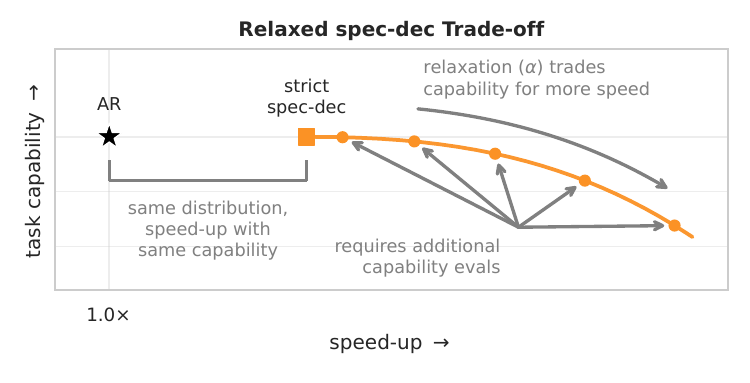}
    \vspace{-3mm}
    \caption{\textbf{Left}: The speed-up model in \cref{eq:speedup-approx}. Drafter relative cost $c_\text{rel}$ has a strong impact on speed-up, the best choice of $N_\text{draft}$, and the speed-up gain from increased average accepted length $\bar{l}_\text{accept}$. \textbf{Right}: \texttt{strict} \texttt{spec-dec} gives lossless speed-up, whilst \texttt{relaxed} \texttt{spec-dec} trades capability for even more speed. If deployment is \textit{capability constrained}, the former is lossless so it doesn't require the burden of additional capability evaluation, whilst the latter does.}
    \label{fig:relaxed-sd-illustration}
\end{figure}

\paragraph{Modelling and understanding speed-up.} Intuitively, we expect speed-ups to be large when 1) the cost of the auxiliary drafter is much lower than the cost of verification, 2) the cost of parallel verification is not much more than a single \texttt{AR} decoding step, and 3) long draft segments are accepted. We will formalise this intuition in this section. The speed-up $S$ achieved by speculative decoding can be quantified as, 
\begin{equation}\label{eq:speedup}
S=\frac{\text{\texttt{spec-dec} token throughput}}{\text{\texttt{AR} token throughput}}=
\frac{\text{time for \texttt{AR} to generate $\bar{l}_\text{accept}+1$ tokens}}{\text{time to draft and verify $N_{\text{draft}}$ tokens}}
\approx
\frac{
    (\bar{l}_{\text{accept}}+1) C_{\text{AR}}
}{
    N_{\text{draft}} C_{\text{drafter}} + C_{\text{verify}}
},
\end{equation}
where $N_{\text{draft}}$ is the draft length, $\bar{l}_{\text{accept}}$ is the average number of accepted draft tokens at $N_\text{draft}$, $C_\text{AR}$ is the time for one \texttt{AR} decoding step of the main LLM $p$, $C_\text{drafter}$ is the time for one decoding step of the auxiliary drafter model $q$,\footnote{We note that drafting doesn't have to be autoregressive, and some recent work proposes one-step parallel drafting \citep{an2026pard,chen2026dflashblockdiffusionflash}. Modelling such approaches would require a simple adjustment to the denominator of \cref{eq:speedup} to $C_{\text{drafter}} + C_{\text{verify}}$. In this work, we will focus on autoregressive drafters as they represent the majority of existing drafters for speculative decoding
.} and $C_\text{verify}$ is the time for one parallel verification step. \cref{eq:speedup} shows that for greater speed-up, verification cost should be minimised, \ie close to the cost of a single original \texttt{AR} step, $C_{\text{verify}}\approx C_{\text{AR}}$. This typically happens, as discussed above, when \texttt{AR} generation is \textit{memory bound} so additional parallel compute has a small effect on inference latency. If the compute burden for each verification step is increased by higher token parallelism, then $C_{\text{verify}}$ may become meaningfully larger than $C_{\text{AR}}$, reducing speed-up. Empirical results support this, with investigations showing that increasing batch size can reduce speed-ups \citep{singh2025acceleratingllminference,liu2026speculativedecodingperformanceillusion}. In the later analysis of this work, we will make the common assumption $C_{\text{verify}}\approx C_{\text{AR}}$, allowing us to define the relative cost $c_\text{rel}=C_{\text{drafter}}/C_\text{AR}$ giving a simplified speed-up model,
\begin{equation}\label{eq:speedup-approx}
S
\approx
\frac{
    \bar{l}_{\text{accept}}+1
}{
    1 + N_{\text{draft}} c_{\text{rel}}
}.
\end{equation}
Average accepted draft length $\bar{l}_{\text{accept}}$ is intuitively higher when the drafter distribution is better aligned (over the draft length) with the verifier $p$.  In fact, the per-token acceptance probability (on average over the drafter distribution),
\begin{equation}\label{eq:tv}
    \Pr(\text{accept draft token})=\mathbb E[a] = 1-\text{TV}(q,p) = 1-\tfrac{1}{2}\sum_x |p(x)-q(x)|,
\end{equation}
is related to the total-variation distance between $q$ and $p$ \citep{spec-dec-theory}.
Considering both \cref{eq:speedup,eq:speedup-approx} we can see that we want to make \textit{design} choices that reduce the cost of the drafter $C_{\text{drafter}}$ and increase the average accepted length $\bar{l}_\text{accept}$. The draft length $N_\text{draft}$ on the other hand is an inference-time hyperparameter, and is subsequently optimised for maximum speed-up after other design choices are selected. It is important to note that the above factors \textit{interact}, complicating the maximisation of speed-up. A more lightweight drafter could reduce both $C_{\text{drafter}}$ and $\bar{l}_\text{accept}$ and may only achieve greater final speed-up $S$ after $N_\text{draft}$ is re-optimised. Simply decreasing drafter cost without changing its distribution $q$, \eg via kernel or hardware optimisations, may change the optimal value of $N_\text{draft}$. This is shown on the left of \cref{fig:relaxed-sd-illustration}. Suppose we have a drafter that achieves $ \bar{l}_{\text{accept}}=2.5$ when $N_{\text{draft}}=3$ and $ \bar{l}_{\text{accept}}=5$ when $N_{\text{draft}}=8$; if we model speed-up according to \cref{eq:speedup-approx} as $c_\text{rel}$ is reduced, the better choice of draft length flips, since it becomes cheaper to draft more tokens, \ie longer $N_\text{draft}$ is ``unlocked''. Moreover, speed-up is sensitive to $c_\text{rel}$, increasingly so as $c_\text{rel}$ is reduced. Suppose we can also improve $\bar{l}_{\text{accept}}$ via better drafter alignment; we see that the (multiplicative) effect is highest when $c_\text{rel}$ is low and $N_\text{draft}$ is high. Moreover, unlocking a longer $N_\text{draft}$ can give greater headroom for $\bar{l}_{\text{accept}}$ to improve. Thus, a practictioner may choose to focus on reducing $c_\text{rel}$ first, and then consider improving $\bar{l}_{\text{accept}}$.
\begin{takeaway}[elevated]{takeaway:speedup-factors}
        There are many factors that contribute to \texttt{spec-dec} speed-up. We want to be \textit{memory bound} so the cost of parallel verification is similar to a single \texttt{AR} step. Drafter $q$ should be low-cost and well-aligned with the LLM $p$ so that a draft can be quickly generated and many draft tokens are accepted. After other design choices have been fixed, the draft length $N_\text{draft}$ should then be optimised. These factors interact and should be considered together, \eg a smaller drafter may draft faster but have more tokens rejected and change the optimal $N_\text{draft}$.
\end{takeaway}
We note the existing research often targets a single aspect (\eg draft acceptance or drafter cost) and then compares speed-up under a limited setting. For example, an increase in $\bar{l}_\text{accept}$ may be measured in a paper for a specific $N_\text{draft}$, however, real-world $C_\text{drafter}$ may lead a practitioner to choose a lower optimal $N_\text{draft}$ where real speed-up improvements are much lower. ``Real-world'' speed-ups may also be measured in settings where neither $c_\text{rel}$ nor $N_\text{draft}$ are well optimised, failing to provide readers with a clear idea of how an approach may translate to realisable improvements. 
For example, \citet{liu2026speculativedecodingperformanceillusion,hao2026cactus} use vLLM to measure speed-up, however, we find that for separate draft models, vLLM has $c_\text{rel}$ much higher than what could be ideally achieved (see \cref{tab:vllm-drafter-cost-diagnostic}). In this paper, we choose to leverage \cref{eq:speedup-approx} to illustrate a more holistic view of speed-up for the reader:
for different methods, we measure $\bar{l}_\text{accept}$ at different draft lengths $N_\text{draft}$ and estimate speed-ups for different relative drafter costs $c_\text{rel}$. As previously discussed, and shown in the left of \cref{fig:relaxed-sd-illustration}, a speed-up model like \cref{eq:speedup-approx} allows a practitioner to understand the impact on speed-up of different optimisations, allowing them to make more informed improvements.
\begin{table}[t]
    \centering
    \small

    \resizebox{\linewidth}{!}{%
    \begin{tabular}{@{}lccccc@{}}
        \toprule
        & $C_\text{AR}$
        & \makecell{\texttt{AR} mode $C_\text{drafter}$}
        & \makecell{\texttt{AR} mode $c_\text{rel}$}
        & \makecell{\texttt{spec-dec} mode $C_\text{drafter}$}
        & \makecell{\texttt{spec-dec} mode $c_\text{rel}$} \\
        \midrule
        Qwen3 0.6B drafter + 32B verifier & $19.53\,\text{ms/token}$ & $5.19\,\text{ms/token}$ & $0.266$ & $16.54\,\text{ms/token}$ & $0.847$ \\
        \bottomrule
    \end{tabular}%
    
    }
        \caption{Inference costs on a single-GPU NVIDIA GH200 Grace Hopper Superchip, using vLLM 0.20.1 and PyTorch 2.11.0+cu130. \texttt{spec-dec} mode is draft token cost for stock vLLM speculative decoding at $N_\text{draft}=5$. vLLM's built-in \texttt{spec-dec} implementation is under-optimised with high $c_\text{rel}$, see \cref{app:vllm} for a discussion.}
            \label{tab:vllm-drafter-cost-diagnostic}
        \vspace{-3mm}
\end{table}
\begin{takeaway}{takeaway:speedup-proxy-model}
    Although research papers may emphasise end-to-end ``real'' speed-ups, these are often measured on a narrow software/hardware setup, which may differ to a practitioner's own. A flexible speed-up proxy model enables practitioners to \textit{reason} over varying deployment scenarios, illuminating bottlenecks and guiding optimisations such as kernel engineering. Real measurements are nevertheless important to validate final behaviour at deployment.
\end{takeaway}

\subsection{Relaxing Speculative Decoding (\texttt{relaxed} \texttt{spec-dec})}
In this paper, we are interested in investigating a recent area of research which argues that the exact preservation of the verifier distribution $p$ is too ``strict'' and that by ``relaxing'' this requirement, advantages can be obtained over \texttt{strict} \texttt{spec-dec}. We narrow our scope to \textit{training-free }approaches that apply a per-token relaxation \textit{away from} $p$/\texttt{strict} \texttt{spec-dec}, to keep comparisons manageable. Such approaches can be applied directly to existing \texttt{spec-dec} setups without the overhead of additional training and avoid the associated risk of distribution drift from the training data. We start by introducing a framework that unifies existing approaches and then discuss some practical caveats of deployment. 

Our framework is laid out in \cref{alg:relaxed-spec-dec}. We introduce \textit{relaxed} target distributions $\pi$ that replace $p$ at the rejection, residual sampling, and bonus sampling steps in the original \texttt{strict} \texttt{spec-dec} algorithm ($\pi^\text{rej},\pi^\text{res},\pi^\text{bonus}$). The relaxed target distributions $\pi[q,p,x,\alpha](\cdot)$ depend on a relaxation parameter $\alpha$ and can potentially depend on the drafter and verifier distributions $p,q$ as well as the drafted token $x$. Given the diverse presentation in the literature, our framework allows us to cleanly describe and compare various existing approaches under a unified setting. 

We note that ``relaxing'' the target distribution is typically motivated by the idea of \textit{trading off} task capability for \textit{generative speed} by accepting more draft tokens than \texttt{strict} \texttt{spec-dec} \citep{narasimhan2025faster,holsman-etal-2025-fuzzy} (we will discuss paper-specific motivations and narratives later). This is illustrated in the right of \cref{fig:relaxed-sd-illustration}. Practically, deployment is either \textit{capability} constrained or \textit{speed} constrained, that is to say one must achieve a target capability or speed. Which type of constraint then determines what \textit{must} be evaluated in order to tune the relaxation parameter $\alpha$ for deployment. Therefore, a practitioner should consider \textit{the cost of evaluation} when deciding whether to choose to deploy relaxed speculative decoding. For example, it can be easy to measure latencies on the fly for a speed-constrained edge application and accordingly adjust $\alpha$, however, the cost of evaluating the capability of a frontier generalist model across all intended deployment scenarios may be considerable \citep{liang2023helm,huggingface2026evalscompute,openai2026swebenchverified}. Crucially, \texttt{strict} \texttt{spec-dec} doesn't require any capability re-evaluation (in theory) as it preserves the distribution of original model $p$.\footnote{There still may be numerical effects that lead to deviations, so small-scale sanity checks may still be necessary.}
\begin{takeaway}[elevated]{takeaway:relaxation-evaluation-cost}
    ``Relaxing'' speculative decoding is typically motivated by the \textit{trading off} of \textit{task capability} for more \textit{generative speed}.
    When deployment is \textit{capability constrained}, no additional capability evaluation is needed for \texttt{strict} \texttt{spec-dec}, as it preserves the verifier distribution. On the other hand, relaxation necessitates re-evaluation (\cref{fig:relaxed-sd-illustration} right).  
    If evaluation is costly, it may not be worth deploying \texttt{relaxed} \texttt{spec-dec} over \texttt{strict} \texttt{spec-dec}.
\end{takeaway}

One important detail that many existing works neglect is that deviating from $p$ via relaxation may change the average generated response length vs \texttt{AR} or \texttt{strict} \texttt{spec-dec}, which may affect overall generation time and thus speed-up. Thus for relaxed speculative decoding speed-up should be estimated using,
\begin{equation}\label{eq:gen-len-speedup}
    S_\text{relax} \approx\frac{\bar{L}_\text{AR}}{\bar{L}_\text{relax}} \frac{
    (\bar{l}_{\text{accept}}+1) C_{\text{AR}}
}{
    N_{\text{draft}} C_{\text{drafter}} + C_{\text{verify}}
}
\approx
\frac{\bar{L}_\text{AR}}{\bar{L}_\text{relax}}
\frac{
    \bar{l}_{\text{accept}}+1
}{
    1 + N_{\text{draft}} c_{\text{rel}}
},
\end{equation}
where $\bar{L}_\text{AR},\bar{L}_\text{relax}$ are the average generation lengths for \texttt{AR} and \texttt{relaxed} \texttt{spec-dec}. Intuitively, if a relaxed distribution $\pi$ tends to generate longer/shorter responses than $p$ then the effective speed-up will decrease/increase.  
\begin{takeaway}{takeaway:generation-length-speedup}
    Since \texttt{relaxed} \texttt{spec-dec} samples from a different distribution to verifier $p$, speed-up should take into account not only draft acceptance/token throughput but also changes in average response length $\bar{L}$ (\cref{eq:gen-len-speedup}).
\end{takeaway}

\section{Related Work}
We discuss literature that is relevant to accelerating speculative decoding, and thus may be of interest to the reader, but exists outside of the scope of this paper's investigation. 
\paragraph{Learned relaxed speculative decoding.} A number of works, motivated by the idea that \texttt{strict} \texttt{spec-dec} may reject draft tokens that are still semantically correct, suggest to \textit{learn} a module that \textit{decides dynamically} to accept/reject draft tokens. Judge Decoding \citep{bachmann2025judge}, inspired by LLM-as-a-Judge, proposes to train an auxiliary classifier on manually annotated data for this purpose, whilst follow-up AutoJudge \citep{garipov2026autojudge} removes manual annotation by automatically verifying downstream correctness. DIVERSED \citep{wang2026diversed} uses reinforcement learning based on task rewards to learn a module that dynamically adjusts a relaxed target distribution.
As mentioned before, these are out of scope for our investigation due to their additional complexity and reliance on an auxiliary training data distribution. 
\paragraph{Accelerating \texttt{strict} \texttt{spec-dec}.} Since it was originally introduced by \citet{spec-dec}, there have been a varied range of approaches proposed to further accelerate \texttt{strict} \texttt{spec-dec}. SpecInfer \citep{specinfer} and SEQUOIA \citep{SEQUOIA} use tree-based drafting to generate multiple drafts and increase acceptance. REST \citep{he-etal-2024-rest} uses text retrieval to propose drafts at a low cost, rather than using a neural network for the drafter. SpecDec++ \citep{huang2025specdecpp} proposes to train an additional module that adaptively varies the draft length during generation. Block Verification \citep{sun2025blockverification} perform rejection jointly on blocks of draft tokens, rather than token by token, increasing draft acceptance. Speculative Speculative Decoding \citep{kumar2026specspec} parallelises drafting and verification by having the drafter precompute continuations for likely verifier outcomes whilst verification is ongoing, thereby hiding drafter latency if the verification outcome is successfully pre-empted. DistillSpec \citep{zhou2024distillspec} and LK Losses \citep{samarin2026lklosses}
train the drafter to better align with the verifier distribution, increasing draft
token acceptance rates.

\paragraph{Dedicated drafter architectures.} There has been strong trend in the community towards researching lightweight dedicated drafter modules that can rapidly propose short drafts. Such approaches aim to minimise drafter cost whilst maintaining decent draft acceptance, leading to strong practical speed-up (see \cref{fig:relaxed-sd-illustration} left). Medusa \citep{cai2024medusa} uses multiple prediction heads for drafting, whilst the EAGLE series \citep{li2024eagle,li2026eagle3} propose to use a shallow auxiliary autoregressive transformer. Recently PARD \citep{an2026pard}, D-Flash \citep{chen2026dflashblockdiffusionflash} and DSpark \citep{cheng2026dspark} explore further acceleration via parallel/semi-autoregressive draft token generation. These are now generally referred to as multi-token-prediction (MTP) modules and are seeing increasing real-world adoption, frequently appearing in LLM releases such as DeepSeek, Qwen, Gemma, Nemotron, and GLM \citep{liu2024deepseekv3technicalreport,cheng2026dspark,qwen35,lacombe2026gemma4mtp,nvidia_nemotron_3_ultra_2026,glm5technicalreport}, as the default (and sometimes only) supported \texttt{spec-dec} drafter within inference frameworks, \eg vLLM, SGLang \citep{kwon2023efficient,zheng2024sglang,schmittulms2025speculators}. By \textit{bundling} such a module together with an LLM release, deployment complexity is reduced as there is no need to manage and serve an entirely separate model.
\section{A Taxonomy of (Training-free) Relaxed Speculative Decoding}\label{sec:relax-methods}

We now provide the reader with an organised summary of a number of relevant approaches in the literature, that aim to relax speculative decoding in a training-free per-token manner. The aim is for the reader to better understand and compare the motivating narratives and concrete algorithms proposed under a unified framework. 
\cref{tab:relaxed-sd-taxonomy} contains a full summary of narrative categorisations and algorithmic details for the different methods investigated in this paper (methods are fully specified when \cref{tab:relaxed-sd-taxonomy} is combined with \cref{alg:relaxed-spec-dec}). 
\paragraph{Bonus token.}
Before introducing the methods we discuss a practical detail that applies to all of them. As mentioned before, when all draft tokens are accepted, an additional bonus token is sampled from $p$. In the relaxed case, it is possible to replace $p$ with a relaxed distribution $\pi^\text{bonus}$ (\cref{alg:strict-spec-dec,alg:relaxed-spec-dec}). This can be underspecified in the \texttt{relaxed} \texttt{spec-dec} literature, where $\pi^\text{rej},\pi^\text{res}$ are generally the main focus.\footnote{For example, for the method \texttt{CACTUS}, \citet{hao2026cactus} do not specify $\pi^\text{bonus}$ in the paper but set it to $p$ in the published code.} In this work, aligning with paper narratives that generally set $p$ as the capability standard, and code implementations when they are publicly available, we set $\pi^\text{bonus} = p$, other than for Speculative Contrastive Decoding (\texttt{spec-cont-dec}) \citep{yuan-etal-2024-spec-cont-dec}, which is the only method that explicitly aims to \textit{improve capability over} $p$ and unambiguously defines $ \pi^\text{bonus}$:
\begin{equation}
    \pi^\text{bonus} = p \quad \text{for all methods other than }\texttt{spec-cont-dec}\text{ where }\quad \pi^\text{bonus} = \pi^\text{rej} =\pi^\text{res}.
\end{equation}
 

\paragraph{CACTUS (\texttt{CACTUS}).} \citet{hao2026cactus}  aim to solve the constrained optimisation problem:
\begin{equation}
\begin{aligned}
    \pi[q,p,x] \in \arg\max_{r \in \Delta^{V-1}} \quad & \min\left\{\tfrac{r(x)}{q(x)},1\right\} \quad \text{s.t.} \quad & D_{\text{KL}}(r\,\|\,p) \le \alpha.
\end{aligned}
\end{equation}
\ie given a draft token $x$, maximise the acceptance probability whilst constraining the reverse KL-divergence of the relaxed target from $p$. \texttt{CACTUS} optimises given the actual draft token $x$, whilst later approaches focus on objectives in expectation over token distributions. A Taylor expansion of the KL-divergence produces the approximate optimum, 
\begin{equation}
    \pi^\text{rej}[q,p,x](v) = \pi^\text{res} = \begin{cases}\gamma_x & \text{if } v=x,\\ \tfrac{1-\gamma_x}{1-p(x)}p(v) & \text{otherwise},\end{cases} \quad \text{ where } \gamma_x=\min\left\{p(x)+\sqrt{2\alpha p(x)(1-p(x))},1\right\},
\end{equation}
where $\pi(x)$ is boosted as a function of $p(x)$ and other vocab probabilities are uniformly reduced. Notably, if $x$ is outside of the verifier distribution's support (\eg top-$p$/$k$ mask) then it is always rejected regardless of $\alpha$, which is a strict behaviour not shared with many of the other approaches. The method thus interpolates between \texttt{strict} \texttt{spec-dec} and verifier-support-based rejection. 
The narrative is to explicitly control/restrict deviation from $p$, whilst maximising speed-up beyond \texttt{strict} \texttt{spec-dec} to provide a good capability-speed trade-off. The main results in \citet{hao2026cactus} suggest close-to-lossless speed-up can be achieved via relaxation. Additionally, some results suggest capability improves, and the authors speculate in the appendix of \citep{hao2026cactus} that this may be due to an ensembling effect. 
\begin{table}[t!]
\vspace{-8mm}
    \centering

    \renewcommand{\arraystretch}{1.25}
    \resizebox{\linewidth}{!}{%
    \begin{tabular}{l|cccc|lcc}
        \toprule
        \taxhead{\textbf{method}} &
        \taxhead{\textbf{trade}\\\textbf{capability}\\\textbf{for more speed}} &
        \taxhead{\textbf{improve}\\\textbf{capability}} &
        \taxhead{\textbf{control}\\\textbf{deviation}\\\textbf{from $p$}} &
        \taxhead{\textbf{rely on}\\\textbf{drafter $q$ for}\\\textbf{generation}} &
        \taxhead{\textbf{rejection}\\\textbf{target dist.}\\$\pi_{\alpha}^{\text{rej}}(q,p,x)(v)$} &
        \taxhead{\textbf{residual}\\\textbf{target dist.}\\$\pi_{\alpha}^{\text{res}}(v)$} &
        \taxhead{\textbf{bonus}\\\textbf{target dist.}\\$\pi_{\alpha}^{\text{bonus}}(v)$} \\
        \midrule
        \begin{tabular}[c]{@{}l@{}}\texttt{CACTUS}\\\citep{hao2026cactus}\end{tabular}
            & \taxcheck & \taxsoftcheck & \taxcheck &
            & \begin{tabular}[c]{@{}l@{}}$\begin{cases}\gamma_x & \text{if } v=x,\\ \tfrac{1-\gamma_x}{1-p(x)}p(v) & \text{otherwise}\end{cases}$\\\taxnote{\gamma_x=\min\{p(x)+\sqrt{2\alpha p(x)(1-p(x))},1\}}\end{tabular}
            & \begin{tabular}[c]{@{}c@{}}same as rej.\end{tabular}
            & \begin{tabular}[c]{@{}c@{}}$p(v)$\end{tabular} \\
        \midrule
        \begin{tabular}[c]{@{}l@{}}\texttt{mentored-dec} $\beta=1$\\\citep{Tran-Thien_2023}\end{tabular}
            & \taxcheck &  & \taxcheck &
            & \begin{tabular}[c]{@{}l@{}}$\min\{p(v)/(1-\alpha),1\}$\end{tabular}
            & \begin{tabular}[c]{@{}c@{}}$p(v)$\end{tabular}
            & \begin{tabular}[c]{@{}c@{}}$p(v)$\end{tabular} \\
        \midrule
        \begin{tabular}[c]{@{}l@{}}\texttt{r-fuzzy}\\\citep{holsman-etal-2025-fuzzy}\end{tabular}
            & \taxcheck &  & \taxcheck & \taxcheck
            & \begin{tabular}[c]{@{}l@{}}$\begin{cases}q(v) & \text{if } \text{Div}(p,q)<\alpha,\\ p(v) & \text{otherwise}\end{cases}$\end{tabular}
            & \begin{tabular}[c]{@{}c@{}}same as rej.\end{tabular}
            & \begin{tabular}[c]{@{}c@{}}$p(v)$\end{tabular} \\
        \midrule
        \begin{tabular}[c]{@{}l@{}}\texttt{spec-casc-opt}\\\citep{narasimhan2025faster}\end{tabular}
            & \taxcheck & \taxsoftcheck &  & \taxcheck
            & \begin{tabular}[c]{@{}l@{}}$\begin{cases}q(v) & \text{if } \max_u q(u)\ge\max_u p(u)-\alpha\operatorname{TV}(p,q),\\ p(v) & \text{otherwise}\end{cases}$\end{tabular}
            & \begin{tabular}[c]{@{}c@{}}same as rej.\end{tabular}
            & \begin{tabular}[c]{@{}c@{}}$p(v)$\end{tabular} \\
        \midrule
        \begin{tabular}[c]{@{}l@{}}\texttt{ens}\\\citep{wang2026diversed}\end{tabular}
            & \taxcheck & \taxsoftcheck & \taxcheck & \taxcheck
            & \begin{tabular}[c]{@{}l@{}}$\alpha q(v) + (1-\alpha)p(v)$\end{tabular}
            & \begin{tabular}[c]{@{}c@{}}same as rej.\end{tabular}
            & \begin{tabular}[c]{@{}c@{}}$p(v)$\end{tabular} \\
        \midrule
        \begin{tabular}[c]{@{}l@{}}\texttt{spec-cont-dec}\\\citep{yuan-etal-2024-spec-cont-dec}\end{tabular}
            &  & \taxcheck &  & 
            & \begin{tabular}[c]{@{}l@{}}$\text{softmax}[z_\pi](v)$\\\taxnote{z_\pi(v)=z_p(v)+\alpha[z_p(v)-z_q(v)]\text{ if }v\in\mathcal M_{\text{top}\text{-}p/k}\text{ otherwise }-\infty}\\\taxnote{\mathcal M_{\text{top}\text{-}p/k}\text{ is the set of logits within the top-}p\text{/top-}k\text{ mask}}\end{tabular}
            & \begin{tabular}[c]{@{}c@{}}same as rej.\end{tabular}
            & \begin{tabular}[c]{@{}c@{}}same as rej.\end{tabular} \\
        \bottomrule
    \end{tabular}%
    }
        \caption{Taxonomy of \texttt{relaxed} \texttt{spec-dec} methods and narratives within our framework from \cref{alg:relaxed-spec-dec}. \taxcheck~indicates when an idea that is explicitly part of the paper narrative, whilst \taxsoftcheck~indicates when an idea is suggested implicitly.}
        \label{tab:relaxed-sd-taxonomy}
        \vspace{-4mm}
\end{table}
\paragraph{Mentored Decoding (\texttt{mentored-dec}).} \citet{Tran-Thien_2023} aims to solve a very similar problem as \texttt{CACTUS},
\begin{equation}\label{eq:cactus}
\begin{aligned}
    \pi[q,p] \in \arg\max_{r \in \Delta^{V-1}} \quad
    & \mathbb{E}_{x\sim q}\left[\min\left\{\tfrac{r(x)}{q(x)},1\right\}\right]
    \quad \text{s.t.} \quad
    & D_{\text{KL}}(p\,\|\,r) \le \alpha.
\end{aligned}
\end{equation}
This differs from \texttt{CACTUS} by optimising expected-over-$q$ rather than draft-token-conditioned acceptance, and by constraining the forward rather than reverse KL from $p$. We adopt the version where $\beta=1$ used in \citet{narasimhan2025faster} (see their Appendix F.6), as it is simpler and shown to perform very similarly to the full version.\footnote{We note that the full version is also more unwieldy to implement as it requires a per-token optimisation over parameters during inference.} This gives,
\begin{equation}\label{eq:mentored}
    \pi^{\text{rej}}[p,\alpha](v)=\min\left\{\tfrac{p(v)}{1-\alpha},1\right\},\quad
    \pi^{\text{res}}(v)=p(v),
\end{equation}
which boosts $\pi^\text{rej}(x)$ and thus acceptance similarly to \texttt{CACTUS} (compare \cref{eq:mentored} with $\gamma_x$ from \cref{eq:cactus}) and then uses $p$ for residual sampling after rejection rather than a relaxed target. The narrative in \citet{Tran-Thien_2023} is also very explicitly about controlling deviation from $p$ and maximising acceptance. Thus, \texttt{mentored-dec} is an algorithm that very much belongs in the same bucket as \texttt{CACTUS} both in terms of story and practical implementation. 
\paragraph{(Reducible) Fuzzy Speculative Decoding (\texttt{r-fuzzy}).} \citet{holsman-etal-2025-fuzzy}  propose to relax the rejection requirement using a divergence criterion between $q$ and $p$ to accept draft tokens if the drafter distribution $q$ is similar to the verifier $p$:
\begin{equation}\label{eq:fuzzy}
    \pi^{\text{rej}}[q,p](v)=\pi^{\text{res}}=
    \begin{cases}
        q(v), & \text{if } \text{Div}(p,q)<\alpha,\\
        p(v), & \text{otherwise}.
    \end{cases}
\end{equation}
We use Jensen-Shannon Divergence as recommended in \citet{holsman-etal-2025-fuzzy}.\footnote{Jensen-Shannon Divergence avoids divide-by-zero issues that may occur with KL-Divergence and non-overlapping support from top-$p$/$k$.} \texttt{r-fuzzy} interpolates between \texttt{strict} \texttt{spec-dec} and always accepting draft tokens from $q$. The story is that the drafter can be relied upon to unlock further speed-up if its distribution is similar to the verifier, with a strong focus on enabling a trade-off between capability and speed, which \texttt{strict} \texttt{spec-dec} cannot provide. Note that plugging the above \cref{eq:fuzzy} into \cref{alg:relaxed-spec-dec} is the ``reducible'' version of the algorithm, whilst \citet{holsman-etal-2025-fuzzy} mainly discuss a simpler \texttt{fuzzy} version that directly chooses between $q$ and $p$ using the divergence criterion. We only investigate the reducible variant (\texttt{r-fuzzy}) as it has \textit{provably better} draft acceptance whilst sampling from the same relaxed target distribution (see \cref{app:r-fuzzy-fuzzy-optimality} for a proof).
\begin{takeaway}{takeaway:r-fuzzy-reducible}
   Reducible fuzzy speculative decoding (\texttt{r-fuzzy}) has provably the same or better expected speed-up than regular fuzzy speculative decoding, whilst sampling tokens from exactly the same target distribution, even though the latter is the main focus in the paper that introduces them \citep{holsman-etal-2025-fuzzy}.
\end{takeaway}
\paragraph{Speculative Cascades (\texttt{spec-casc-opt}).} \citet{narasimhan2025faster} combine ideas from model cascades with speculative decoding. Cascades/early-exit networks \citep{huang2018multiscale,Xia2023window} are a well-established approach for increasing the inference efficiency of neural networks. The idea is to run a set of models sequentially, only running later models (at greater cost) when necessary, \ie exiting computation on a sample early if a satisfactory result has already been achieved. Speculative decoding comes with a drafter+verifier model pair, so \citet{narasimhan2025faster} propose to only sample from the verifier distribution $p$ (and incur the cost of potential rejection) if a criterion is met. For binary decision $d\in\{0,1\}$, where $d=1$ means deferring to $p$, their \texttt{spec-casc-opt} rule optimises the per-token loss,
\begin{equation}
    \mathcal L_{\text{spec}}(d;q,p)=
    \mathbb E_{v\sim p_\star}\!\left[(1-d)\ell(v,q)+d\left\{\ell(v,p)+\alpha\operatorname{TV}(p,q)\right\}\right],
\end{equation}
where $p_\star$ is the true next-token data distribution, $\ell$ is the  $0\text{-}1$ loss, and $\alpha\text{TV}$ is the expected cost of deferring to the verifier under rejection as in \cref{alg:relaxed-spec-dec}. Approximating expected losses under $q$ and $p$ with $1-\max_u q(u)$ and $1-\max_u p(u)$ gives deferral rule $d^\star(q,p)=\mathbf{1}\{\max_u q(u)<\max_u p(u)-\alpha\operatorname{TV}(p,q)\}$ and relaxed distribution,
\begin{equation}
    \pi^{\text{rej}}[q,p,\alpha](v)=\pi^{\text{res}}=
    \begin{cases}
        q(v), & \text{if } \max_u q(u)\ge\max_u p(u)-\alpha\operatorname{TV}(p,q),\\
        p(v), & \text{otherwise}.
    \end{cases}
\end{equation}
Like \texttt{r-fuzzy}, this approach interpolates between \texttt{strict} \texttt{spec-dec} and always accepting draft tokens from $q$.\footnote{Although \cite{narasimhan2025faster} define $\alpha$ as a non-negative cost of deferral, in practice, setting a negative $\alpha$ is completely possibly. It simply encourages a preference towards the verifier, and pushing $\alpha\rightarrow-\infty$ recovers \texttt{strict} \texttt{spec-dec}.} The narrative is not only to rely on $q$, but to use the confidence of $q$ \textit{itself} to make the judgment, and to \textit{prefer} $q$ over $p$ since deferring incurs a cost. As such it is the approach that leans the hardest into trusting the drafter model $q$. \citet{narasimhan2025faster} generally are able to produce close-to-lossless speed-ups, and also suggest that relying on the drafter in some cases can improve capability, showing in some experiments that capability can improve with relaxation.\footnote{We note that \citet{narasimhan2025faster} propose another approach \texttt{Tok} which we omit from the main body, see \cref{app:tok} for a discussion.} 
\paragraph{Ensemble (\texttt{ens}).} \citet{wang2026diversed} introduce a simple weighted drafter+verifier ensemble as a baseline,
\begin{equation}\label{eq:ens}
    \pi^\text{rej}[q,p](v) = \pi^\text{res} = \alpha q(v) + (1-\alpha)p(v),
\end{equation}
and show that it is \textit{pareto-optimal} for trading off deviation from $p$ measured by $\operatorname{TV}(\pi,p)$ against expected rejection rate given by $\operatorname{TV}(q,\pi)$ (\cref{eq:tv}). They motivate their work by illustrating how \texttt{strict} \texttt{spec-dec} may reject semantically acceptable draft tokens and suggest that the drafter can be relied on more. Their experiments show ensembling as in \cref{eq:ens} may improve capability over $p$. We omit \citet{wang2026diversed}'s main approach DIVERSED as out of scope, as it requires learning a model to predict $\alpha$ on each task. 
\paragraph{Speculative Contrastive Decoding  (\texttt{spec-cont-dec}).} \citet{yuan-etal-2024-spec-cont-dec}\footnote{\citet{fu2025collab} also discuss the same idea in a contemporaneous publication} observe that the drafter-verifier structure of speculative decoding is naturally exploitable for contrastive decoding \citep{li-etal-2023-contrastive,obrien2024contrastive,chang-etal-2024-explaining}, where the logits $z$ of a weak ``amateur'' model are subtracted from an ``expert'', \textit{improving} the task capability of the ``expert''. By substituting the drafter as the amateur, contrastive decoding can be performed ``for free'' potentially leading to both speedup over \texttt{AR} and better capability. Contrastive decoding is usually performed with a ``plausibility set'' of logits, in order to prevent noisy lower-valued logits degrading performance \citep{yuan-etal-2024-spec-cont-dec,chang-etal-2024-explaining}. 
We use the set of logits within the top-$p$/top-$k$ mask \citep{fan-etal-2018-hierarchical,holtzman2020curious} of the verifier $p$, $\mathcal M_{\text{top}\text{-}p/k}$, as it is already present in our experiments, avoiding the need to introduce additional hyperparameters, 
\begin{equation}\label{eq:contrast}
    \!\pi^\text{rej}[q,p](v)\!=\!\pi^\text{res}\!=\!\pi^\text{bonus}\!=\!\text{softmax}[z_\pi](v),\quad\! z_\pi(v)=z_p(v)+\alpha[z_p(v)\!- \!z_q(v)]\text{ if }v\in\mathcal M_{\text{top}\text{-}p/k}\text{ otherwise }\!-\!\infty.
\end{equation}
As noted earlier, unlike the above approaches, \texttt{spec-cont-dec} uses the relaxed target distribution for the all-accept bonus token, as the relaxed distribution is meant to have superior capability compared to $p$. It is the only approach whose narrative is focused on improving capability, whilst also benefiting from the speed-up of speculative decoding. 
\paragraph{Implementation cost.}
We find the implementation overhead of relaxed approaches to be roughly the same as \texttt{strict} \texttt{spec-dec}, incurring a slight memory capacity cost increase. A brief analysis in \cref{app:mem-cost} on Qwen 3.5 shows that the additional overhead becomes notable for small models (\eg 2B) with large vocabulary sizes at larger batch sizes (\eg 32), but is negligible for larger models like Qwen 3.5 27B. Readers should keep their own use case in mind.

\section{Experimental Evaluation}
We now present a comparative evaluation of the relaxed speculative decoding methods just discussed. The aim of our experiments is to fairly compare the approaches, to test the validity of the various narratives discussed in \cref{sec:relax-methods,tab:relaxed-sd-taxonomy}, and to provide the reader with empirical \textcolor{GraphcoreFindingTitle}{Finding}s that may be of use for their own work. We aim to make up for a few shortcomings found in the evaluations of the existing literature. As many of the approaches in \cref{sec:relax-methods} have been published recently and contemporaneously, they often only compare against a subset of the other methods. As previously mentioned in \cref{sec:spec-dec}, existing works generally present narrow evaluations in terms of $N_\text{draft}$ and real-world drafter cost $C_\text{drafter}$, and they often neglect the impact of overall generation length on speed-up.
Experiments are often performed on out-of-date benchmarks such as CNN/Daily Mail \citep{see-etal-2017-cnndm} and XSum \citep{narayan-etal-2018-xsum} that may not accurately reflect real-world applications.
Finally, the impact of the language modelling ability of the drafter is rarely explored, and the methods in \cref{sec:relax-methods} have yet to be evaluated with the increasingly popular and relevant approach of a bundled dedicated multi-token-prediction (MTP) drafter module.

To this end we measure average accepted length $\bar{l}_\text{accept}$ over a range of draft lengths $N_\text{draft}\in \{3,5,10,20\}$ and relaxation values $\alpha$ and then use \cref{eq:gen-len-speedup}, which takes into account generation length, with different potential drafter/verifier relative time cost $c_\text{rel}\in \{0.05,0.2,0.5\}$ to estimate speed-up over a broad range of settings. We consider 3 drafter+verifier pairs, \{Qwen3.5 MTP+27B, Qwen3 0.6B+32B, Qwen3 8B+32B\}, that respectively represent a dedicated single-layer MTP drafter module, a weak language model drafter and a strong language model drafter. For our main results we consider AIME 2024 (\texttt{AIME 24}), a challenging maths olympiad dataset that is widely used in model evaluations \citep{aime24}.
We also evaluate on GPQA Diamond (\texttt{GPQA}) and LiveCodeBench Lite v6 (\texttt{LCB}) \citep{rein2024gpqa,jain2025livecodebench}. \{\texttt{AIME24}, \texttt{GPQA}, \texttt{LCB}\} are modern challenging benchmarks that test a model's reasoning ability, for which we use \texttt{thinking mode}. To reduce measurement noise from stochastic sampling, we repeat dataset score measurements for each evaluation point \{drafter, verifier, $N_\text{draft}$, \texttt{method}, $\alpha$\} so that the order of $10^3$ queries are made to the model. We use standard Qwen stochastic sampling parameters (top-$p$/$k$) as softmax support truncation is now standard practice for LLM generation \citep{noarov2026toppkused}. To test another model family we also evaluate Llama 3.2 1B+3.1 70B \citep{llama3herd} on \texttt{GPQA}. See \cref{app:exp-deets} for full experimental details (\eg $\alpha$ values, evaluation harness \etc).


\subsection{Empirical Results and Findings}
\begin{figure}[p]
    \centering
    \vspace{-10mm}
    \includegraphics[width=\linewidth]{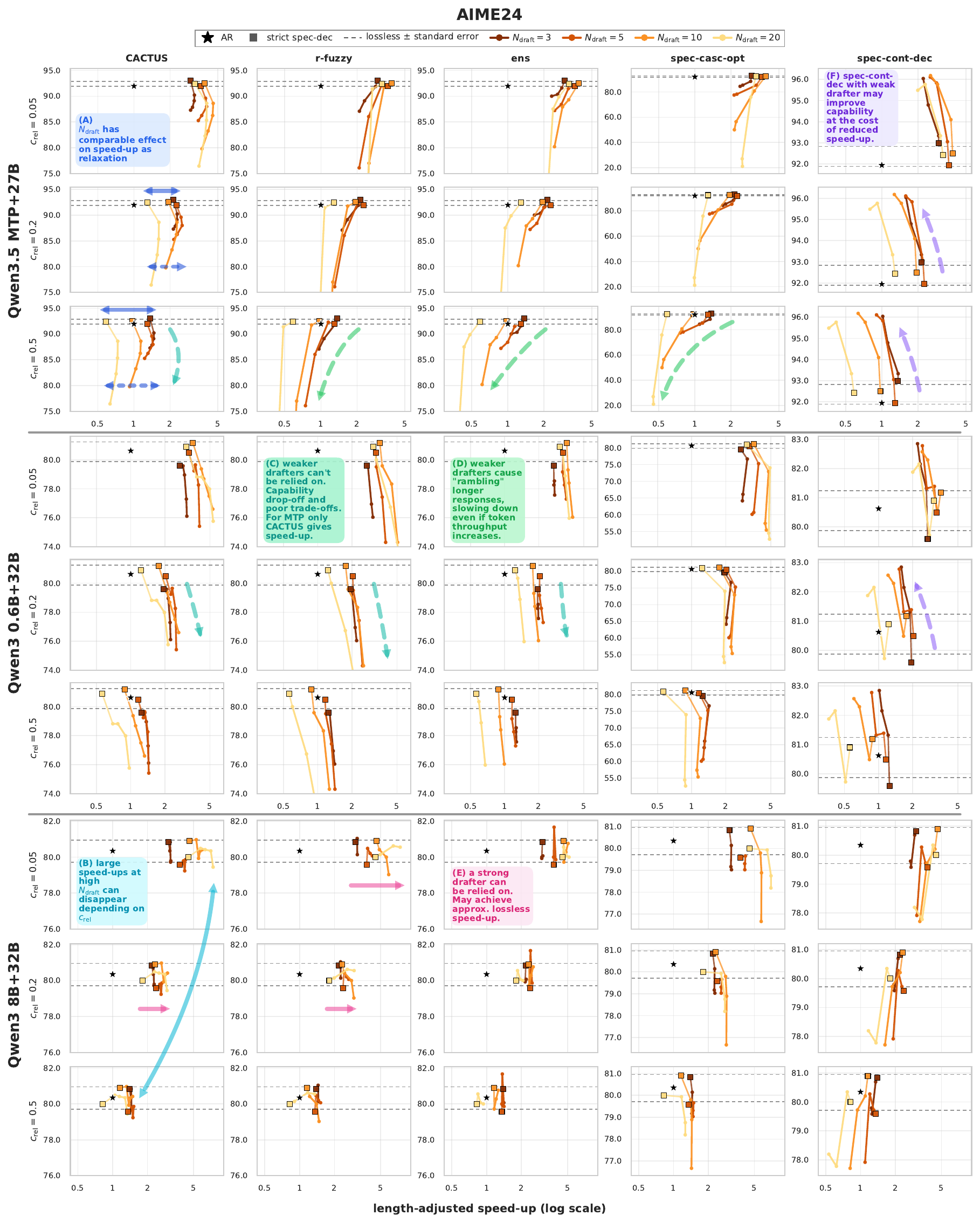}
    \caption{Capability-speed trade-offs on \texttt{AIME24} for \texttt{relaxed} \texttt{spec-dec} methods across drafter+verifier pairs, relative drafter costs $c_\text{rel}$ and draft lengths $N_\text{draft}$. Standard error lines indicate measurement noise from stochastic sampling. \textcolor{SupergridAnnA}{\textbf{(A)}} $N_\text{draft}$ has a comparable effect on speed-up as relaxation. \textcolor{SupergridAnnB}{\textbf{(B)}} large speed-ups/increases in average accepted length $\bar{l}_\text{accept}$ at high $N_\text{draft}$ may become irrelevant depending on $c_\text{rel}$, if the best $N_\text{draft}$ is shorter. \textcolor{SupergridAnnC}{\textbf{(C)}} lightweight drafters can't be relied on for relaxation, with worse trade-offs and clear capability drop-offs -- for MTP only \texttt{CACTUS} gives speed-up. \textcolor{SupergridAnnD}{\textbf{(D)}} lightweight drafters cause ``rambling'' longer responses, leading to \textit{slowdown}, even if token throughput increases. \textcolor{SupergridAnnE}{\textbf{(E)}} a stronger drafter language model \textit{can} be relied on for relaxation, potentially giving approx. lossless speed-up, but it will incur higher $c_\text{rel}$ in reality limiting speed-up. \textcolor{SupergridAnnF}{\textbf{(F)}} \texttt{spec-cont-dec} with a weak drafter language model may \textit{improve} capability at the cost of reduced speed-up.}
    \label{fig:aime-qwen-mtp-0p6b-8b-relaxed-supergrid}
\end{figure}

\cref{fig:aime-qwen-mtp-0p6b-8b-relaxed-supergrid} shows our main results, a grid of capability-speed trade-offs over \texttt{methods} and drafter+verifier pairs on \texttt{AIME24}, with highlighted findings. There is variance in capability evaluation due to random sampling of responses, which can be seen in the differences between lossless configurations -- standard error lines indicate measurement noise. The reader should be aware when interpreting the results that lightweight drafters will incur lower $c_\text{rel}$ in reality, even if all drafters are evaluated over the same $c_\text{rel}$ grid. Full results for \texttt{AIME24}, \texttt{GPQA},  \texttt{LCB} and Llama3 are in \cref{app:additional-results}. We will leave discussion of \texttt{spec-cont-dec} to the end as it is distinct from the other approaches.
\begin{figure}[t]
\vspace{-8mm}
    \centering
    \includegraphics[width=\linewidth]{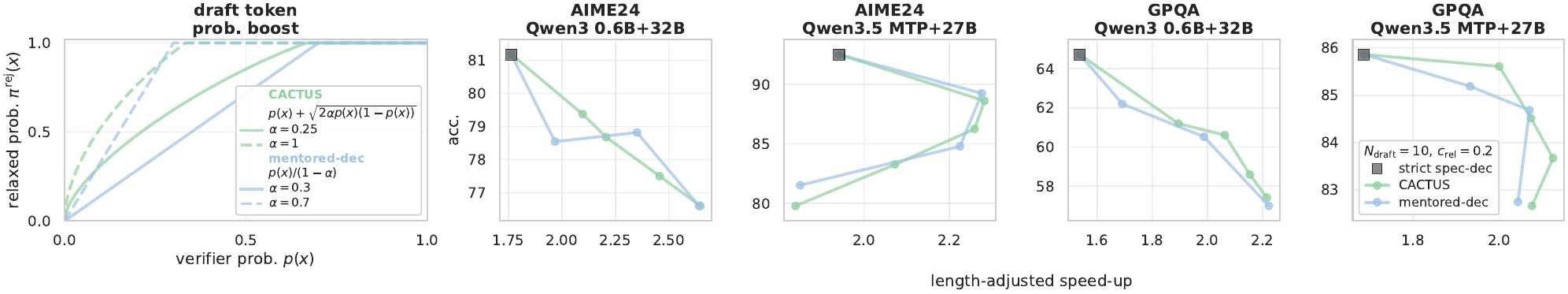}
    \vspace{-3mm}
    \caption{Comparison between \texttt{CACTUS} and \texttt{mentored-dec}. Left: a visualisation of how the relaxed probability of the draft token $x$ is boosted, right: empirical trade-off comparisons. Their algorithmic and narrative similarities (\cref{sec:relax-methods}) result in similar empirical behaviour. We omit \texttt{mentored-dec} in favour of \texttt{CACTUS} in the main paper body for brevity. }
    \label{fig:cactus-vs-mentored-d10-c0p2}. 
    \vspace{-5mm}
\end{figure}
\paragraph{The similarity of \texttt{CACTUS} and \texttt{mentored-dec}.} 
As discussed in \cref{sec:relax-methods}, \texttt{mentored-dec} \citep{Tran-Thien_2023} and the more recently published \texttt{CACTUS} \citep{hao2026cactus} are in fact quite similar algorithmically. The left of \cref{fig:cactus-vs-mentored-d10-c0p2} shows that the probability boost given to draft token $x$ is very similar. We observe generally that both methods behave similarly empirically, shown in the right of \cref{fig:cactus-vs-mentored-d10-c0p2}, and so we omit \texttt{mentored-dec} from the main body results for brevity.  

\begin{finding}{finding:cactus-mentored-similar}
    Mentored speculative decoding \citep{Tran-Thien_2023} and CACTUS \citep{hao2026cactus} are narratively and algorithmically very similar and correspondingly perform similarly in experiments (\cref{fig:cactus-vs-mentored-d10-c0p2}). 
\end{finding}

\paragraph{The impact of draft length.} \cref{fig:aime-qwen-mtp-0p6b-8b-relaxed-supergrid} shows that varying draft length has a considerable impact on achievable speed-up, having comparable or greater impact than relaxation, whilst not requiring any additional capability evaluation.  It also demonstrates how the best draft length changes with $c_\text{rel}$, demonstrating the need to evaluate over \{$N_\text{draft}$,$c_\text{rel}$\} \ie showing relaxation gives a very large speed-up over \texttt{strict} \texttt{spec-dec} at $N_\text{draft}=20$ can be misleading if, for a practitioner's $c_\text{rel}$, $N_\text{draft}=5$ gives higher overall speed-up and much smaller improvements over \texttt{strict} \texttt{spec-dec}. 
\begin{finding}[elevated]{finding:draft-length-priority}
    Optimising the draft length $N_\text{draft}$ with \texttt{strict} \texttt{spec-dec} has a comparable impact on speed-up to relaxation, whilst not requiring additional capability evaluation and retaining the lossless guarantee. This should be prioritised first. The relative drafter cost $c_\text{rel}$ has a large impact on the best $N_\text{draft}$ and overall speed-up. (\cref{fig:aime-qwen-mtp-0p6b-8b-relaxed-supergrid} \textcolor{SupergridAnnA}{\textbf{(A)}})
\end{finding}
Intuitively, a lower optimal $N_\text{draft}$ may impose a lower ceiling on achievable improvement from relaxation. \cref{tab:qwen-aime-cactus-accept-span} shows $\bar{l}_\text{accept}$ for different $N_\text{draft}$, where there is clearly much more headroom for improvement at longer draft lengths. 
\begin{finding}{finding:draft-length-headroom}
    A lower optimal $N_\text{draft}$ can limit the achievable possible additional speed-up from relaxation, if $\bar{l}_\text{accept}$ is already close to $N_\text{draft}$ for \texttt{strict} \texttt{spec-dec}, diminishing the impact of relaxation. (\cref{fig:aime-qwen-mtp-0p6b-8b-relaxed-supergrid} \textcolor{SupergridAnnB}{\textbf{(B)}}, \cref{tab:qwen-aime-cactus-accept-span})
\end{finding}

\begin{table}[t]
    \vspace{-10mm}
    \centering
    \setlength{\tabcolsep}{2.3pt}

    \resizebox{\linewidth}{!}{%
    \begin{tabular}{l*{20}{c}}
        \toprule
        & \multicolumn{5}{c}{$N_\text{draft}=3$}
        & \multicolumn{5}{c}{$N_\text{draft}=5$}
        & \multicolumn{5}{c}{$N_\text{draft}=10$}
        & \multicolumn{5}{c}{$N_\text{draft}=20$} \\
        \cmidrule(lr){2-6}\cmidrule(lr){7-11}\cmidrule(lr){12-16}\cmidrule(lr){17-21}
        Drafter+verifier & strict & $\alpha=0.1$ & $0.25$ & $1$ & $10$ & strict & $\alpha=0.1$ & $0.25$ & $1$ & $10$ & strict & $\alpha=0.1$ & $0.25$ & $1$ & $10$ & strict & $\alpha=0.1$ & $0.25$ & $1$ & $10$ \\
        \midrule
        Qwen3.5 MTP+27B & 2.44 & 2.74 & 2.76 & 2.82 & 2.84 & 3.52 & 4.16 & 4.21 & 4.36 & 4.46 & 4.82 & 6.34 & 6.43 & 7.07 & 7.50 & 5.36 & 7.73 & 7.86 & 9.24 & 10.68 \\
        Qwen3 0.6B+32B & 2.14 & 2.31 & 2.37 & 2.46 & 2.54 & 3.05 & 3.39 & 3.52 & 3.74 & 3.90 & 4.32 & 5.12 & 5.42 & 6.02 & 6.52 & 5.25 & 6.54 & 7.15 & 8.28 & 9.36 \\
        Qwen3 8B+32B & 2.46 & 2.64 & 2.70 & 2.78 & 2.83 & 3.72 & 4.14 & 4.29 & 4.47 & 4.58 & 5.94 & 7.11 & 7.55 & 8.16 & 8.55 & 8.15 & 10.75 & 11.95 & 13.74 & 14.95 \\
        \bottomrule
    \end{tabular}%
    }
        \caption{Mean accepted draft tokens $\bar{l}_\text{accept}(\leq N_\text{draft})$ on \texttt{AIME24} for \texttt{CACTUS}. Lower $N_\text{draft}$ limits achievable improvements from relaxation. The strong drafter is able to better take advantage of longer $N_\text{draft}$.}
    \label{tab:qwen-aime-cactus-accept-span}
\end{table}

\begin{figure}[t]
\vspace{-4mm}
    \centering
    \includegraphics[width=\linewidth]{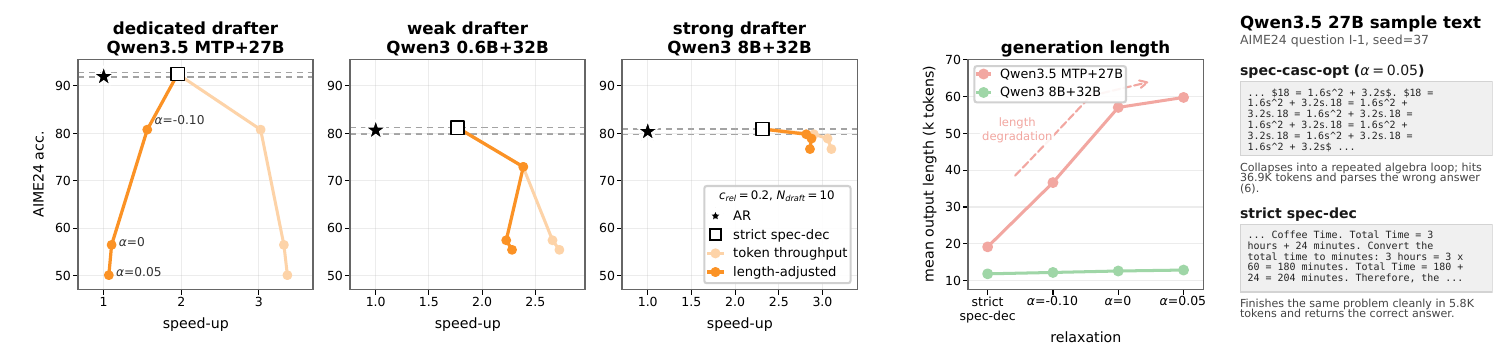}
    \vspace{-6mm}
    \caption{\textbf{Left}: (\texttt{spec-casc-opt}) drafters that are weak language models improve token throughput much more than \textit{length-adjusted} speed-up. \textbf{Right}: relaxation into an MTP drafter increases generation length with rambling responses.}
    \label{fig:weak-strong-drafter-takeaway}
    \vspace{-3mm}
\end{figure}
\paragraph{The impact of drafter capability on relaxation.} As previously noted, the community is increasingly moving towards dedicated lightweight MTP drafter modules that are \textit{specialised for drafting} rather than \textit{general language modelling}. Moreover, many open serving frameworks \textit{only} support MTP for \texttt{spec-dec} if it is bundled with the model, \ie the user can be \textit{forced} to use MTP rather than freely choose a drafter (see \cref{app:mtp-support} for a discussion). We find that such MTP modules are naturally unsuited to \texttt{relaxed} \texttt{spec-dec}, precisely due to the fact that they are weak general language models, and so \textit{can't be relied on for relaxation}, even if they are \textit{decent drafters under strict rejection}. \cref{fig:aime-qwen-mtp-0p6b-8b-relaxed-supergrid} \textcolor{SupergridAnnC}{\textbf{(C)}} shows that for Qwen3.5 MTP+27B, only \texttt{CACTUS}/\texttt{mentored-dec}, which tightly control deviation from $p$ rather than relying on $q$, are able to produce a useful capability-speed trade-off. Methods that explicitly rely on $q$ (\eg \texttt{spec-casc-opt}, see \cref{tab:relaxed-sd-taxonomy}) fail with capability drop-offs \textit{without speed-ups} (discussed later). Using a small general language model (Qwen3 0.6B+32B) gives more useful trade-offs than MTP across evaluated methods, but still shows consistent clear capability degradation, a far cry from the near-lossless capability demonstrated in many \texttt{relaxed} \texttt{spec-dec} papers.
\begin{finding}[elevated]{finding:weak-drafters-tradeoffs}
    A good drafter under \textit{strict} rejection $\neq$ a reliable language model for relaxed rejection. Drafters $q$ that are \textit{weaker standalone language models} give poor capability-speed trade-offs with clear capability degradation for \texttt{relaxed spec-dec}. In particular, the dedicated MTP drafter struggles to produce useful trade-offs across most methods, with the exception of $\texttt{CACTUS}/\texttt{mentored-dec}$ that tightly control deviation from $p$. This is notable as MTP drafting may be the \textit{only supported method} for \texttt{spec-dec} for a given model+serving framework. (\cref{fig:aime-qwen-mtp-0p6b-8b-relaxed-supergrid} \textcolor{SupergridAnnC}{\textbf{(C)}}) 
\end{finding}
\cref{fig:aime-qwen-mtp-0p6b-8b-relaxed-supergrid} \textcolor{SupergridAnnD}{\textbf{(D)}} actually shows \textit{slowdowns} vs \texttt{strict} \texttt{spec-dec} in many cases, even though relaxation should increase $\bar{l}_\text{accept}$ and token throughput. This is because relaxing into a drafter that is a weak language model may cause the model to take longer to reach a ``thinking'' conclusion, and can even induce generation pathologies (\eg rambling, repetition loops) that lead to longer generations and overall response times. This reduces speed-up compared to what token throughput or $\bar{l}_\text{accept}$ would indicate, highlighting the need to consider generation length (\cref{eq:gen-len-speedup}). This is shown in \cref{fig:weak-strong-drafter-takeaway}.\footnote{Interestingly, $\alpha=0$ is the theoretical minimum suggested by \citep{narasimhan2025faster} for \texttt{spec-casc-opt}, as it represents having no ``cost'' of deferring to the verifier. However, in our experiments, 0 is clearly too high when the drafter is unreliable.} 
\begin{finding}{finding:weak-drafters-slowdowns}
    Relaxing with drafters that are weak language models may lead to \textit{longer} ``rambling'' responses, reducing speed-up, potentially to less than \texttt{strict} \texttt{spec-dec}, even if token throughput increases.
    (\cref{fig:aime-qwen-mtp-0p6b-8b-relaxed-supergrid} \textcolor{SupergridAnnD}{\textbf{(D)}}, \cref{fig:weak-strong-drafter-takeaway}) 
\end{finding}
\cref{fig:aime-qwen-mtp-0p6b-8b-relaxed-supergrid} \textcolor{SupergridAnnE}{\textbf{(E)}} actually shows, on the other hand, that the strong drafter (Qwen3 8B+32B) is able to provide much better (sometimes lossless up to measurement noise) trade-offs with relaxation, aligning much closer with results presented in the literature. This is intuitive as a drafter that is a stronger standalone model matches the narrative that $q$ can be relied on more than \texttt{strict} \texttt{spec-dec} allows, and methods with this story (\cref{tab:relaxed-sd-taxonomy}) in particular see improvements compared to results with drafters that are weak language models. 
Moreover, a stronger drafter also intuitively unlocks greater acceleration via longer $N_\text{draft}$, as the drafter is able to generate longer spans of acceptable text (\cref{tab:qwen-aime-cactus-accept-span}). We note the reader should keep in mind that a stronger drafter will incur considerably higher $c_\text{rel}$ in reality than a lightweight drafter, so realisable speed-up will be reduced in comparison (and \cref{fig:aime-qwen-mtp-0p6b-8b-relaxed-supergrid} should be interpreted with this in mind).
\begin{finding}[elevated]{finding:strong-drafters-tradeoffs}
    A drafter that is a stronger language model enables better (sometimes close-to-lossless) trade-offs for \texttt{relaxed} \texttt{spec-dec}, in particular, methods that explicitly rely on $q$ (\cref{tab:relaxed-sd-taxonomy}) improve considerably vs weaker drafters. However, achievable speed-ups will be limited by large drafter cost, and the community is moving towards dedicated MTP modules rather than drafters that are individually strong standalone language models. (\cref{fig:aime-qwen-mtp-0p6b-8b-relaxed-supergrid} \textcolor{SupergridAnnE}{\textbf{(E)}})
\end{finding}
This highlights a \textit{conflict of narratives} where many \texttt{relaxed} \texttt{spec-dec} approaches predicate their methods on the idea that the drafter is a decent language model which can afford relaxation, whilst the community shift towards dedicated MTP drafters \textit{leans into }the guarantees of \texttt{strict} \texttt{spec-dec} as a crux to disregard general language modelling and focus on faster drafting, whilst only needing to maintain short-range $\bar{l}_\text{accept}$ under strict rejection. \cref{fig:aime-qwen-mtp-0p6b-8b-relaxed-supergrid,tab:qwen-aime-cactus-accept-span} show that the MTP drafter even has higher $\texttt{strict}$ speed-up than Qwen3 0.6B, but worse relaxed trade-offs. As such, although existing methods for \texttt{relaxed} \texttt{spec-dec} are not strictly \textit{invalid}, they are \textit{unsuited} for many mainstream drafters. 
\paragraph{\texttt{spec-cont-dec} can improve capability at the cost of speed-up.} The narrative of this approach, and thus its empirical behaviour, is different to other \texttt{relaxed} \texttt{spec-dec} methods. \cref{fig:aime-qwen-mtp-0p6b-8b-relaxed-supergrid} shows that \texttt{spec-cont-dec} is able to \textit{trade-off speed-up for better capability}. It meaningfully improves on \texttt{AIME24} when the drafter is weak, matching the original intuition that pushing the model distribution \textit{away from an amateur} can improve capability \citep{li-etal-2023-contrastive}. This also intuitively explains why speed-up is reduced, as \cref{eq:contrast} pushes $\pi$ away from $q$ increasing $\operatorname{TV}(q,\pi)$ (\cref{eq:tv}). In the case of a strong drafter, we note a degradation in capability, which also aligns with the same intuition, as we wouldn't expect a non-amateur drafter to produce a useful contrastive signal. We note that with a weak drafter, capability improvement does not happen in every case (\eg Qwen 3.5 MTP+27B \texttt{GPQA} in \cref{fig:gpqa-qwen-mtp-0p6b-8b-relaxed-supergrid}), although in our experiments we do not observe capability degradation beyond measurement noise. This is in line with previous results \citep{obrien2024contrastive}, that also demonstrate non-universal improvements. Other relaxation approaches that may have been suggested to improve capability (\cref{tab:relaxed-sd-taxonomy}) do not demonstrate this in our experiments. 
\begin{finding}[elevated]{finding:contrastive-capability-speed}
    \texttt{spec-cont-dec} is empirically shown to offer an \textit{alternative} trade-off where generative speed-up can be sacrificed for (potentially) better task capability. This is intuitive as the target distribution is pushed \textit{away} from the drafter distribution, thereby reducing acceptance. In the case of speculative inference, it can be implemented with little to no extra cost, as the drafter can be readily used as the amateur model. Other approaches, that may have been suggested to improve capability (see \cref{tab:relaxed-sd-taxonomy}), do not do so in our experiments.
\end{finding}



\begin{figure}[t]
\vspace{-6mm}
    \centering
    \includegraphics[width=\linewidth]{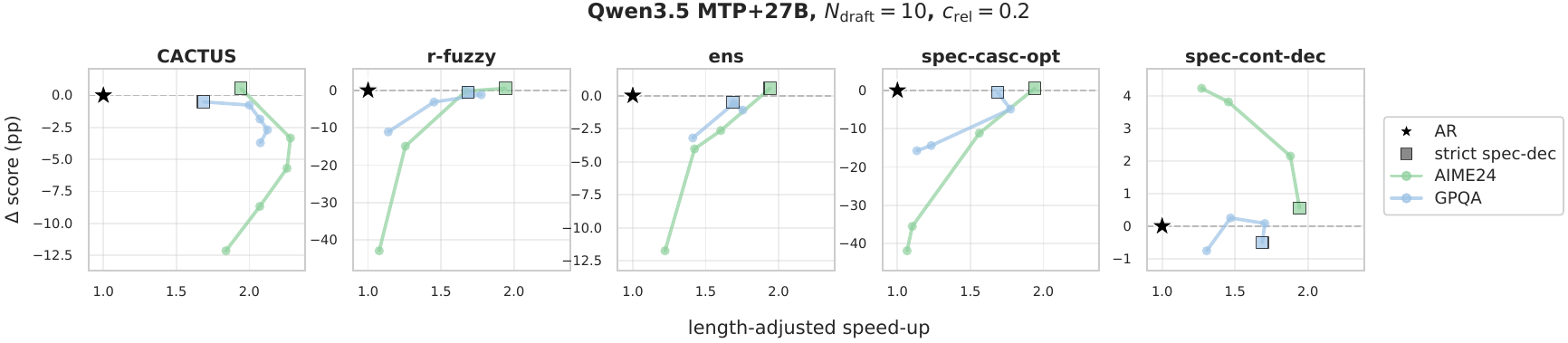}\vspace{-3mm}
    \caption{Relaxation hyperparameters do not reliably generalise between evaluation tasks. Comparing \texttt{AIME24} and \texttt{GPQA} with the same relaxation values, it is clear that the calibration of $\alpha$ is considerably different, \eg the smallest relaxation of $\alpha=0.1$ for \texttt{CACTUS} leads to considerably more capability loss in percentage points on \texttt{AIME24} than \texttt{GPQA}.}
    \label{fig:hyperparameter-generalisation-aime-gpqa-qwen35-mtp}\vspace{-3mm}
\end{figure}

\paragraph{The generalisation of relaxation parameter $\alpha$ over tasks.} There are no expectations or guarantees that relaxation parameter $\alpha$ will generalise between tasks, and we find experimentally that the calibration of $\alpha$ can be quite different between tasks. \cref{fig:hyperparameter-generalisation-aime-gpqa-qwen35-mtp} shows that the same values of $\alpha$ (see \cref{app:samp-params} for values) induce markedly different capability-speed trade-offs between \texttt{AIME24} and \texttt{GPQA}, with the former's accuracy being much more sensitive to $\alpha$ than the latter's.
\begin{finding}[elevated]{finding:alpha-task-generalization}
    Relaxation hyperparameters are not guaranteed to generalise between evaluation tasks, and we show they empirically fail to do so. Thus, for reliable multi-task deployment (that is task-capability constrained), evaluation for relaxation parameter tuning should be across \textit{all tasks}, and will incur the corresponding cost.
\end{finding}


\section{Concluding remarks}
In this work, we performed a practical investigation of recent approaches to  \textit{relaxed} speculative decoding. Such methods deviate from the strict guarantee of matching the verifier distribution in pursuit of additional acceleration, controlled capability-speed trade-offs, or potential capability gains. Our investigation suggests that the achievable results are highly setting-dependent. Draft length and relative drafter cost can affect speed-up as much, if not more than, relaxation; weak or dedicated MTP drafters are not reliable language models for relaxed rejection; stronger drafters can potentially give approximately lossless relaxed speed-up; 
and relaxation parameters do not reliably transfer across tasks. Relaxation can be useful, especially with stronger drafters or methods that tightly control deviation from the verifier, but it should be treated by practitioners as a deployment-specific optimisation potentially requiring further capability evaluation and tuning, rather than a drop-in replacement for strict standard speculative decoding.

\bibliography{main}
\bibliographystyle{tmlr}

\clearpage
\appendix

\startcontents[appendices]
\section*{Appendix Contents}

{
\hypersetup{linkcolor=black}
\setcounter{tocdepth}{2}
\printcontents[appendices]{}{1}{\setcounter{tocdepth}{2}}
}
\clearpage

\section{Glossary of Notation}\label[appendix]{app:gloss}

We summarise the main notations used in the paper in \cref{tab:glossary}.
{
\renewcommand{\arraystretch}{1.25}

\begin{table}[h!]
    \centering

    \resizebox{.8\textwidth}{!}{
    \colorbox{gray!5}{
    \begin{tabular}{ll}
        \toprule
        \textbf{Notation} & \textbf{Meaning} \\
        \midrule
        $q,p$ & Drafter and verifier models; $p$ is the LLM being accelerated \\
        $q(\cdot),p(\cdot)$ & Drafter/verifier next-token distributions \\
        $q_t,p_t,q_{t+i},p_{t+i}$ & Next-token distributions at positions $t$ or $t+i$ \\
        $x_{<t}$ & Autoregressive context before position $t$ \\
        $x_t,x_{t+i}$ & Tokens at positions $t$ and $t+i$ \\
        $x,v,V$ & Draft token, vocabulary index, and vocabulary size \\
        $\Delta^{V-1}$ & Probability simplex over the vocabulary \\
        $N_\text{draft}$ & Number of draft tokens proposed per step \\
        $i,i_\text{end}$ & Draft offset and first rejected/end position \\
        $a_i,\operatorname{Ber}(\cdot)$ & Acceptance indicator and Bernoulli distribution \\
        $\operatorname{norm}(\cdot)$ & Normalisation to a valid probability distribution \\
        $p^\text{res}(\cdot)$ & Actual residual distribution sampled after rejection \\
        $p^\text{bonus}(\cdot)$ & Actual bonus-token distribution sampled after full acceptance \\
        $S,S_\text{relax}$ & Token-throughput and response-length-adjusted speed-up \\
        $\bar{l}_\text{accept}$ & Average number of accepted draft tokens \\
        $\bar{L}_\text{AR},\bar{L}_\text{relax}$ & Average response lengths under \texttt{AR} and \texttt{relaxed} \texttt{spec-dec} \\
        $C_\text{AR},C_\text{drafter},C_\text{verify}$ & Time costs for one AR verifier step, drafter step, and verification step \\
        $c_\text{rel}$ & Relative drafter time cost, $C_\text{drafter}/C_\text{AR}$ \\
        $\text{TV},D_{\text{KL}},\operatorname{Div}$ & Distribution distances/divergence criteria \\
        $\pi$ & Generic relaxed target next-token distribution \\
        $\pi^\text{rej},\pi^\text{res},\pi^\text{bonus}$ & Relaxed targets for rejection, residual construction, and bonus sampling \\
        $\pi^{\text{rej}}_{t+i},\pi^{\text{res}}_{t+i},\pi^{\text{bonus}}_{t+N_\text{draft}}$ & Position-specific relaxed targets \\
        $\pi^\text{rej}[q,p,x,\alpha](v)$ & Rejection target after method-specific construction at token $v$ \\
        $\alpha$ & Relaxation parameter; meaning is method-dependent \\
        $r\in\Delta^{V-1}$ & Candidate relaxed target in optimisation objectives \\
        $\gamma_x$ & \texttt{CACTUS} boosted probability assigned to draft token $x$ \\
        $\beta$ & Mentored-decoding parameter; fixed to $\beta=1$ here \\
        $z_p,z_q,z_\pi$ & Verifier, drafter, and relaxed logits \\
        $\mathcal M_{\text{top}\text{-}p/k}$ & Verifier top-$p$/top-$k$ mask \\
        $d\in\{0,1\}$ & Binary deferral decision in \texttt{spec-casc-opt} \\
        $p_\star,\ell$ & True next-token distribution and $0$-$1$ loss in \texttt{spec-casc-opt} \\
        $\mathcal L_\text{spec}(d;q,p),d^\star(q,p)$ & Speculative-cascades loss and approximate deferral rule \\
        \midrule
        $A_\alpha,\eta_\alpha$ & Trusted top set and mass scaling in appendix \texttt{spec-casc-tok} \\
        $\rho$ & Generic target distribution in the appendix correction proof \\
        $R_t,\pi_t^{\text{\texttt{fuzzy}}}$ & Appendix fuzzy predicate and its target distribution \\
        $K_{\text{\texttt{fuzzy}}},K_{\text{\texttt{r-fuzzy}}}$ & Appendix one-step kernels for \texttt{fuzzy} and \texttt{r-fuzzy} \\
        $A_{\text{\texttt{fuzzy}}},A_{\text{\texttt{r-fuzzy}}}$ & Appendix draft-token acceptance events \\
        \texttt{pass@1} & Appendix code-generation first-sample success rate \\
        \bottomrule
    \end{tabular}
    }
    }
        \caption{Glossary of Notation}
    \label{tab:glossary}
\end{table}
}
\newpage

\section{Speculative Cascades \texttt{Tok}}\label[appendix]{app:tok}
This approach in \citet{narasimhan2025faster} creates a target distribution $\pi$ by keeping only the vocabulary elements of $q$ within a trusted top set of $p$ and then adding a scaled down version of $p$ on top,
\begin{equation}\label{eq:spec-casc-tok}
\begin{aligned}
    \pi^\text{rej}[q,p](v)=\pi^\text{res}(v)
    &=
    \begin{cases}
        q(v)+\eta_\alpha p(v), & \text{if } v\in A_\alpha,\\
        \eta_\alpha p(v), & \text{otherwise},
    \end{cases},\qquad \pi^\text{bonus}(v)=p(v),\\
    A_\alpha&=\{u:p(u)\ge(1-\alpha)\max_w p(w)\},\qquad
    \eta_\alpha=1-\sum_{u\in A_\alpha}q(u).
\end{aligned}
\end{equation}
We omit it from the main body for brevity and because it distorts the target distribution away from $p$, so that minimal relaxation, in this case $\alpha=0$, is not $p$ (it is $p$ for all other approaches), but instead still a combination of $q$ and $p$ based on the $\arg \max$ of $p$. Such top-set truncation may overlap with already present temperature, top-$p$ and top-$k$ adjustments, confusing results.\footnote{This is noted in the reviews on openreview \url{https://openreview.net/forum?id=vo9t20wsmd}.} We do report results for this approach in \cref{app:additional-results}. We find that \textit{minimum} relaxation $\alpha=0$ generally incurs a capability drop if the drafter is a weak language model, with no option to reduce relaxation further. We also find that for drafter models that are weak language models, trade-offs are generally roughly the same or a bit worse than \texttt{CACTUS}/\texttt{mentored-dec} and better than approaches that explicitly rely on $q$. For the strong drafter trade-offs are roughly in line with other approaches, although slightly odd behaviour on \texttt{GPQA} could possibly be attributed to the top-set-distorting behaviour of \cref{eq:spec-casc-tok}.
\section{\texorpdfstring{\texttt{fuzzy} and \texttt{r-fuzzy}}{fuzzy and r-fuzzy}}
\label[appendix]{app:r-fuzzy-fuzzy-optimality}

The following is a one-step conditional statement under exact sampling. It abstracts away implementation details such
as finite block scheduling, caching, numerical approximation, and non-exact residual sampling.
For finite draft blocks, the comparison assumes that both variants use the same bonus-token distribution after a full draft is accepted; in our experiments this is $p_t$.
\begin{result}[\texttt{r-fuzzy} preserves the \texttt{fuzzy} output distribution and weakly improves acceptance]
\label{result:r-fuzzy-fuzzy-kernel}
Fix a prefix $x_{<t}$ and write
\[
q_t(\cdot)=q(\cdot|x_{<t}), \qquad p_t(\cdot)=p(\cdot|x_{<t}).
\]
Let $x_t\sim q_t$ be the draft token, and let
\[
R_t \equiv \operatorname{Div}(p_t,q_t)<\alpha
\]
be the \texttt{fuzzy} relaxation predicate. Define the one-token \texttt{fuzzy} target from \citet{holsman-etal-2025-fuzzy}
\[
\pi^{\text{\texttt{fuzzy}}}_t(v)
=
\begin{cases}
q_t(v), & R_t,\\
p_t(v), & \text{otherwise}.
\end{cases}
\]
Consider \texttt{r-fuzzy} as \cref{alg:relaxed-spec-dec} with
\[
\pi^{\mathrm{rej}}[q_t,p_t,x_t]
=
\pi^{\mathrm{res}}[q_t,p_t,x_t]
=
\pi^{\text{\texttt{fuzzy}}}_t,
\qquad
\pi^{\mathrm{bonus}}[q_t,p_t,x_t]
=
p_t.
\]
Let $K_{\text{\texttt{fuzzy}}}(\cdot\mid x_{<t})$ and
$K_{\text{\texttt{r-fuzzy}}}(\cdot\mid x_{<t})$ denote the one-step
output distributions over the next emitted token given prefix $x_{<t}$.
Let $A_{\text{\texttt{fuzzy}}}$ and $A_{\text{\texttt{r-fuzzy}}}$ denote
the corresponding events that the drafted token $x_t$ is accepted.
Then, under exact residual sampling,
\[
K_{\text{\texttt{r-fuzzy}}}(\cdot\mid x_{<t})
=
K_{\text{\texttt{fuzzy}}}(\cdot\mid x_{<t})
=
\pi^{\text{\texttt{fuzzy}}}_t(\cdot).
\]
Consequently, if the same one-token rule is applied after each realised prefix, the two algorithms generate the same distribution over full sequences. They differ only in how often a draft token is kept rather than resampled.
Moreover,
\[
\Pr(A_{\text{\texttt{r-fuzzy}}}\mid x_{<t})
\geq
\Pr(A_{\text{\texttt{fuzzy}}}\mid x_{<t}),
\]
and, for any prefix $x_{<t}$ such that $R_t$ fails,
\[
\Pr(A_{\text{\texttt{r-fuzzy}}}\mid x_{<t})
=
\sum_v \min\{p_t(v),q_t(v)\}
=
1-\operatorname{TV}(p_t,q_t),
\]
which is the maximal exact acceptance probability for proposal $q_t$ and target $p_t$.
\end{result}

\begin{proof}
We condition throughout on the realised prefix $x_{<t}$, so $q_t$, $p_t$, and $R_t$ are fixed. We use the standard
one-step speculative-correction fact: for proposal $q$ and target $\rho$, accepting a proposed token $v\sim q$ with
probability
\[
\min\{1,\rho(v)/q(v)\}
\]
and, on rejection, sampling from
\[
\operatorname{norm}\big((\rho-q)_+\big)
\]
emits a token with marginal distribution $\rho$. Its acceptance probability is
\[
\sum_v q(v)\min\{1,\rho(v)/q(v)\}
=
\sum_v \min\{\rho(v),q(v)\}
=
1-\operatorname{TV}(\rho,q),
\]
and this is optimal among exact one-step accept/reject corrections with proposal $q$ and target $\rho$
\citep{spec-dec-theory}. As usual, the acceptance ratio is only evaluated on the support of $q$.

If $R_t$ holds, then $\pi^{\text{\texttt{fuzzy}}}_t=q_t$. The \texttt{fuzzy} sampler emits the draft token
$x_t\sim q_t$. The \texttt{r-fuzzy} sampler has the same relaxed target $q_t$, so its acceptance probability is one
and its output distribution is also $q_t$. Thus both kernels equal $\pi^{\text{\texttt{fuzzy}}}_t$ on this branch, and
both draft-token acceptance probabilities are one.

If $R_t$ fails, then $\pi^{\text{\texttt{fuzzy}}}_t=p_t$. The \texttt{fuzzy} sampler rejects the draft token and samples
directly from $p_t$, so its one-step kernel is $p_t$ and its draft-token acceptance probability is zero. The
\texttt{r-fuzzy} sampler is exactly the standard speculative correction with proposal $q_t$ and target $p_t$. By the
fact above, its one-step kernel is also $p_t$, and its draft-token acceptance probability is
\[
\sum_v \min\{p_t(v),q_t(v)\}
=
1-\operatorname{TV}(p_t,q_t),
\]
which is the maximal exact acceptance probability for this proposal and target.

Combining the two cases,
\[
K_{\text{\texttt{r-fuzzy}}}(\cdot\mid x_{<t})
=
K_{\text{\texttt{fuzzy}}}(\cdot\mid x_{<t})
=
\pi^{\text{\texttt{fuzzy}}}_t(\cdot),
\]
and
\[
\Pr(A_{\text{\texttt{r-fuzzy}}}\mid x_{<t})
=
\begin{cases}
1, & R_t,\\
1-\operatorname{TV}(p_t,q_t), & \text{otherwise},
\end{cases}
\qquad
\Pr(A_{\text{\texttt{fuzzy}}}\mid x_{<t})
=
\begin{cases}
1, & R_t,\\
0, & \text{otherwise}.
\end{cases}
\]
Hence \texttt{r-fuzzy} weakly improves the draft-token acceptance probability. Equality of the one-step transition
kernels at every realised prefix implies equality of the induced autoregressive sequence distributions by the chain
rule.
\end{proof}

\section{Experimental Details}\label[appendix]{app:exp-deets}
We evaluate training-free relaxed \texttt{spec-dec} methods over
model--task settings, and report speed with the response-length-aware proxy in
\cref{eq:gen-len-speedup}. Due to the stochastic nature of sampling we repeat each dataset a number of times to reduce measurement noise for each data point (\eg visualised in \cref{fig:aime-qwen-mtp-0p6b-8b-relaxed-supergrid}). Each repeat uses a distinct random seed; for a fixed
model setting and dataset, the same seed set is used across methods, relaxation
parameters, and draft lengths. 

\subsection{Datasets}

Example boxes show rendered system and user message contents before
model-specific chat-template serialisation. They are not exact tokeniser
strings; literal prompt strings and requested output strings are typeset in
monospace.

\paragraph{AIME 2024 (\texttt{AIME24}).}
\texttt{AIME24} \citep{aime24} evaluates olympiad-style contest mathematics
with a three-digit integer answer. The exam is part of the U.S. mathematical
olympiad qualification pathway, which gives context for the level of contest
problem solving being evaluated. We use the 30-question AIME 2024 set
distributed by HuggingFaceH4, with 48 final repeats per problem for each reported
condition; each plotted data point therefore uses 1,440 model queries
($30\times48$). The prompt asks for step-by-step
reasoning and a boxed final answer. We strip Qwen thinking traces before
parsing, then use boxed or final-answer integers; the primary metric is exact
match, with a post-thinking exact-match diagnostic also recorded.\footnote{Dataset/prompt
source: \url{https://huggingface.co/datasets/HuggingFaceH4/aime_2024}.}
\begin{examplebox}
\textbf{User.} Solve the following math problem efficiently and clearly. The last
line of your response should be of the following format:
\texttt{\textquotesingle{}Therefore, the final answer is: \$\textbackslash{}boxed\{ANSWER\}\$. I hope it is correct\textquotesingle{}}
(without quotes) where ANSWER is just the final number or expression that solves
the problem. Think step by step before answering.
\promptquestion{Every morning Aya goes for a $9$-kilometer-long walk and stops at
a coffee shop afterwards. When she walks at a constant speed of $s$ kilometers
per hour, the walk takes her 4 hours, including $t$ minutes spent in the coffee
shop. When she walks $s+2$ kilometers per hour, the walk takes her 2 hours and
24 minutes, including $t$ minutes spent in the coffee shop. Suppose Aya walks at
$s+\frac{1}{2}$ kilometers per hour. Find the number of minutes the walk takes
her, including the $t$ minutes spent in the coffee shop.}
\end{examplebox}

\paragraph{GPQA Diamond (\texttt{GPQA}).}
\texttt{GPQA} \citep{rein2024gpqa} evaluates graduate-level STEM scientific
reasoning in a multiple-choice format. The questions are written for domain
experts in biology, physics and chemistry, so the task is meant to probe
expert-level scientific knowledge rather than general STEM familiarity. We use
the 198-question Diamond split
after applying a fixed answer-choice shuffle following the EleutherAI evaluation
protocol, with 6 final repeats per problem for each reported condition; each
plotted data point therefore uses 1,188 model queries ($198\times6$). The
prompt is zero-shot chain-of-thought with a forced final line, \texttt{The answer
is (X).} We strip Qwen thinking traces, then parse boxed or final-answer choices
before falling back to the last parenthesised choice. Following GPQA usage
conditions, we redact the example question and answer choices here.\footnote{Dataset: \url{https://huggingface.co/datasets/Idavidrein/gpqa}. Prompt/scoring reference: \url{https://github.com/EleutherAI/lm-evaluation-harness/tree/v0.4.11/lm_eval/tasks/gpqa}.}
\begin{examplebox}
\textbf{System.} Here are some example questions from experts. Answer the
question by selecting one of the provided choices.

\textbf{User.}
\promptquestion{[REDACTED]}
\textbf{Choices.} [REDACTED]\par
\textbf{Instruction.} Let's think step by step. After reasoning, end with exactly:
\texttt{The answer is (X).}
\end{examplebox}

\paragraph{LiveCodeBench Lite v6 (\texttt{LCB}).}
\texttt{LCB} tests contest-style code generation from contemporary programming
problems. The benchmark draws from competitive-programming platforms and scores
submissions by execution against tests, giving a difficult contest-programming
style evaluation. We use LiveCodeBench Code Generation Lite v6, the 175-problem v6 test
split and non-cumulative v6 slice of Code Generation Lite
\citep{jain2025livecodebench,livecodebench_lite_v6}, with 6 final repeats per
problem for each reported condition; each plotted data point therefore uses
1,050 model queries ($175\times6$). The chat prompt gives the problem statement, asks for a Python
programme using standard input/output, and requires fenced code. The parser
extracts the final fenced Python block, matching the LiveCodeBench extraction
policy; we execute extracted Python solutions against the bundled LCB-lite tests and report pass@1.\footnote{Dataset: \url{https://huggingface.co/datasets/livecodebench/code_generation_lite}. Benchmark reference: \url{https://github.com/LiveCodeBench/LiveCodeBench}.}
\begin{examplebox}
\textbf{System.} You are an expert Python programmer. You will be given a question
(problem specification) and will generate a correct Python program that matches
the specification and passes all tests.

\textbf{User.} \texttt{\#\#\# Question:}
\promptquestion{Among the 81 integers in the 9-by-9 multiplication table, find
the sum of those that are not $X$. Each cell in row $i$ and column $j$ contains
$i\times j$. The input is a single integer $X$ between 1 and 81. Print the sum,
counting repeated table entries separately.}
\texttt{\#\#\# Format:} Read the inputs from \texttt{stdin} solve the problem and
write the answer to \texttt{stdout} (do not directly test on the sample inputs).
Enclose your code within delimiters as follows. Ensure that when the python
program runs, it reads the inputs, runs the algorithm and writes output to
STDOUT.

\texttt{\textasciigrave\textasciigrave\textasciigrave python}\par
\texttt{\# YOUR CODE HERE}\par
\texttt{\textasciigrave\textasciigrave\textasciigrave}\par
\texttt{\#\#\# Answer: (use the provided format with backticks)}
\end{examplebox}

\subsection{Models, Sampling Parameters and Relaxation Parameters}\label{app:samp-params}

\paragraph{Models.}
We evaluate three reported model settings. The first two use Qwen3-32B as
verifier with either Qwen3-0.6B or Qwen3-8B as drafter \citep{qwen3}. The third
uses Qwen3.5-27B in native-MTP mode, with no separate draft model
\citep{qwen35}.\footnote{Model cards: \url{https://huggingface.co/Qwen/Qwen3-32B}, \url{https://huggingface.co/Qwen/Qwen3-0.6B}, \url{https://huggingface.co/Qwen/Qwen3-8B}, and \url{https://huggingface.co/Qwen/Qwen3.5-27B}.}
All three settings run \texttt{AIME24}, \texttt{GPQA}, and \texttt{LCB}.
As an additional model-family check, we evaluate Llama-3.1-70B-Instruct as
verifier with Llama-3.2-1B-Instruct as drafter on \texttt{GPQA} only
\citep{llama3herd}.\footnote{Model cards: \url{https://huggingface.co/meta-llama/Llama-3.1-70B-Instruct} and \url{https://huggingface.co/meta-llama/Llama-3.2-1B-Instruct}.}

\paragraph{Sampling parameters.}
For all reported Qwen settings, we use the sampling
parameters recommended in the Qwen3 tech report \citep{qwen3} for thinking mode, temperature=0.6, top-$p$=0.95 and top-$k$=20, omitting additional
logit-penalty parameters for simplicity. The standard thinking budget is 32,768 tokens and the maximum
generation length is 36,864 tokens. The Qwen3.5-27B native-MTP setting uses a
65,536-token thinking budget and 69,632-token maximum generation length on
\texttt{AIME24} and \texttt{LCB}.

The generation protocol enforces the Qwen thinking budget operationally.
Generation proceeds up to the thinking budget. If no \texttt{</think>} token
sequence appears, the reasoning block is closed with a fixed early-stop message
and \texttt{</think>}, after which answer generation continues within the
remaining token budget.

For the Llama \texttt{GPQA} run, we use the Hugging Face generation-config sampling
parameters, temperature=0.6 and top-$p$=0.9, with no top-$k$ filtering. Llama does not
use a thinking mode; the maximum generation length is 4096 tokens.

\paragraph{Relaxation parameters.}
For the reported capability-speed grids, all speculative rows use draft lengths
$N_\text{draft}\in\{3,5,10,20\}$. We include \texttt{AR} verifier
baselines, \texttt{strict} \texttt{spec-dec} \citep{spec-dec},
\texttt{CACTUS} \citep{hao2026cactus}, \texttt{ens}
\citep{wang2026diversed}, \texttt{spec-casc-opt} and \texttt{spec-casc-tok}
\citep{narasimhan2025faster}, \texttt{spec-cont-dec}
\citep{yuan-etal-2024-spec-cont-dec}, \texttt{mentored-dec}
\citep{Tran-Thien_2023}, and \texttt{r-fuzzy}
\citep{holsman-etal-2025-fuzzy}. The relaxed parameter grids are
\texttt{CACTUS} $\alpha\in\{0.10,0.25,1.00,10.00\}$, \texttt{ens} drafter
weight $\alpha\in\{0.05,0.10,0.20\}$ (equivalently verifier weight $1-\alpha \in\{0.95,0.90,0.80\}$), \texttt{spec-casc-opt}
$\alpha\in\{-0.10,0.00,0.05\}$, \texttt{spec-casc-tok}
$\alpha\in\{0.00,0.35,0.80\}$, \texttt{mentored-dec}
$\alpha\in\{0.30,0.70,0.95\}$, \texttt{spec-cont-dec}
$\alpha\in\{0.10,0.50,1.00\}$, and \texttt{r-fuzzy} threshold
$\{0.05,0.20,0.30\}$. The Qwen3.5 native-MTP setting uses the same method
definitions with its built-in MTP proposer. The Llama \texttt{GPQA} setting uses
the same draft lengths and relaxation grids (other than $\alpha=0.1$ is unevaluated for \texttt{CACTUS}).

\subsection{Compute Environment}
The main generation experiments were run as single-GPU jobs, on NVIDIA H100
and B200 GPUs, using vLLM 0.20.1 and PyTorch 2.11.0+cu130. We did not use tensor parallelism, data parallelism,
pipeline parallelism, or FSDP within a generation job.  As described in
the main body, speed-up values in the figures are reported with the
response-length-aware proxy in \cref{eq:gen-len-speedup}, rather than as direct
end-to-end wall-clock serving benchmarks.

\section{Additional Results}
\label[appendix]{app:additional-results}

\begin{figure}[p]
    \centering
    \includegraphics[width=\linewidth]{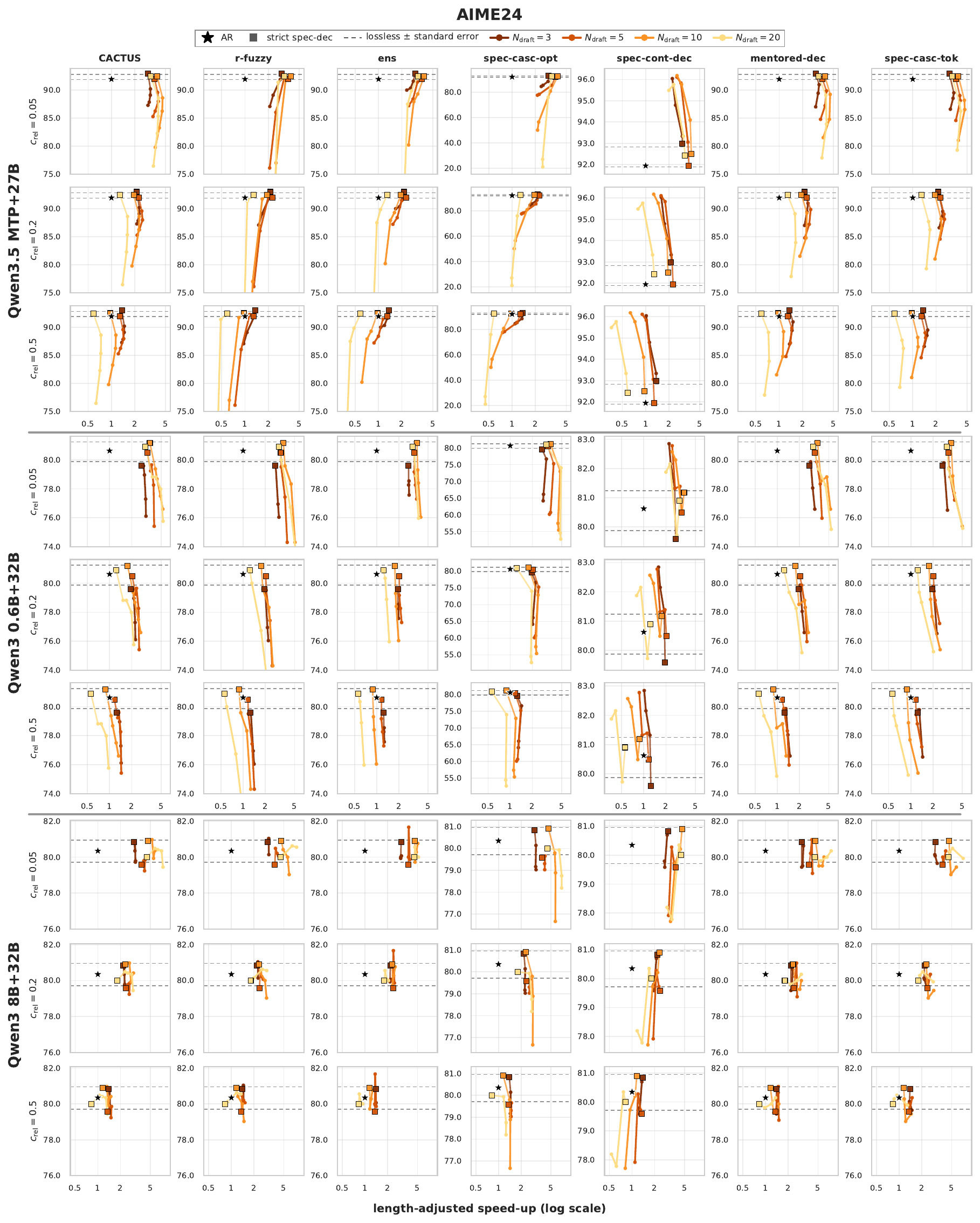}
    \caption{Capability-speed trade-offs on \texttt{AIME24} for \texttt{relaxed} \texttt{spec-dec} methods across drafter+verifier pairs, relative drafter costs $c_\text{rel}$ and draft lengths $N_\text{draft}$. Each block shows one drafter+verifier pair, with method columns and $c_\text{rel}$ rows. The x-axis is length-adjusted speed-up (log scale) and the y-axis is task capability.}
    \label{fig:app-aime-qwen-mtp-0p6b-8b-relaxed-supergrid}
\end{figure}

\begin{figure}[p]
    \centering
    \includegraphics[width=\linewidth]{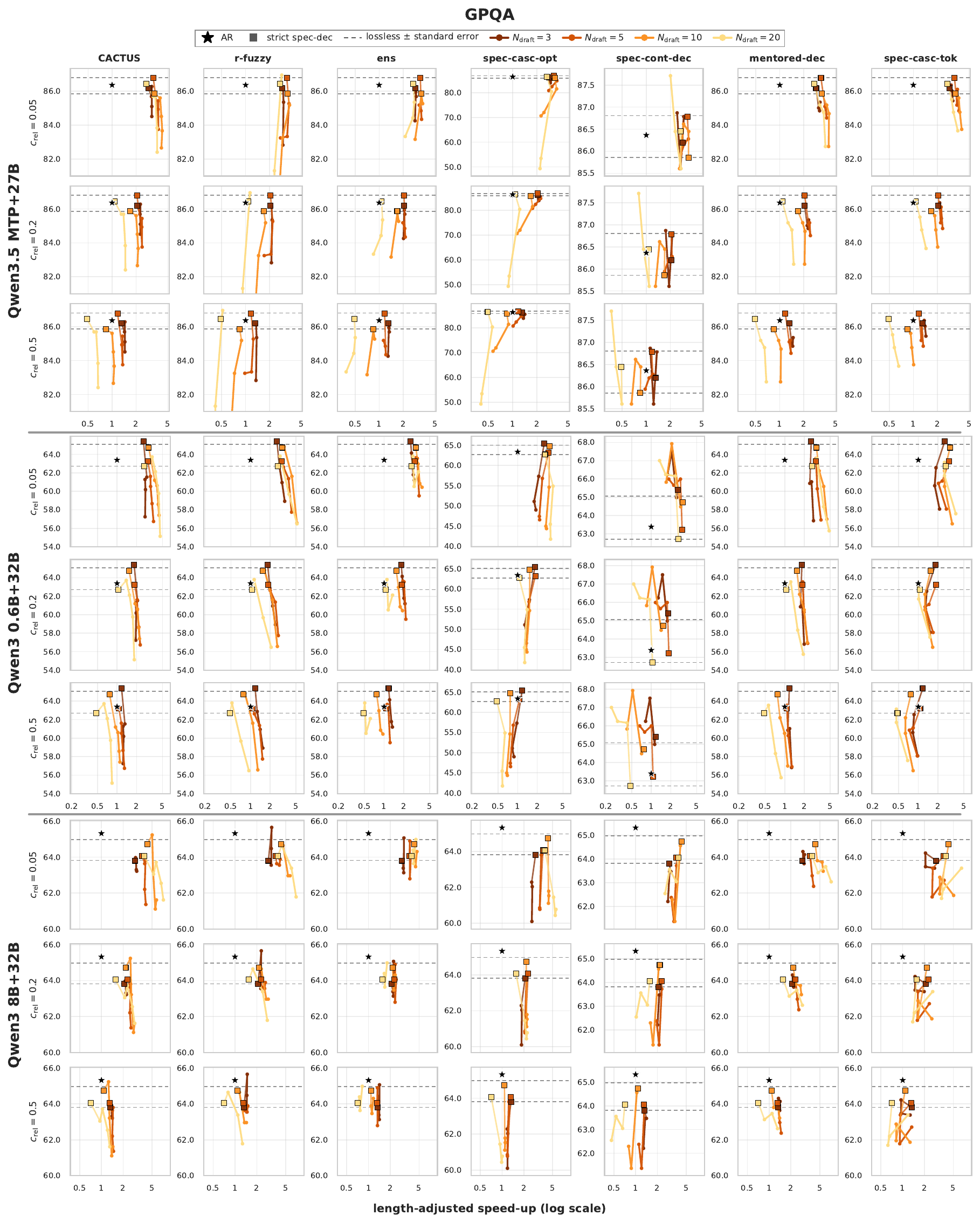}
    \caption{Capability-speed trade-offs on \texttt{GPQA} for \texttt{relaxed} \texttt{spec-dec} methods across drafter+verifier pairs, relative drafter costs $c_\text{rel}$ and draft lengths $N_\text{draft}$. Each block shows one drafter+verifier pair, with method columns and $c_\text{rel}$ rows. The x-axis is length-adjusted speed-up (log scale) and the y-axis is task capability.}
    \label{fig:gpqa-qwen-mtp-0p6b-8b-relaxed-supergrid}
\end{figure}

\begin{figure}[p]
    \centering
    \includegraphics[width=\linewidth]{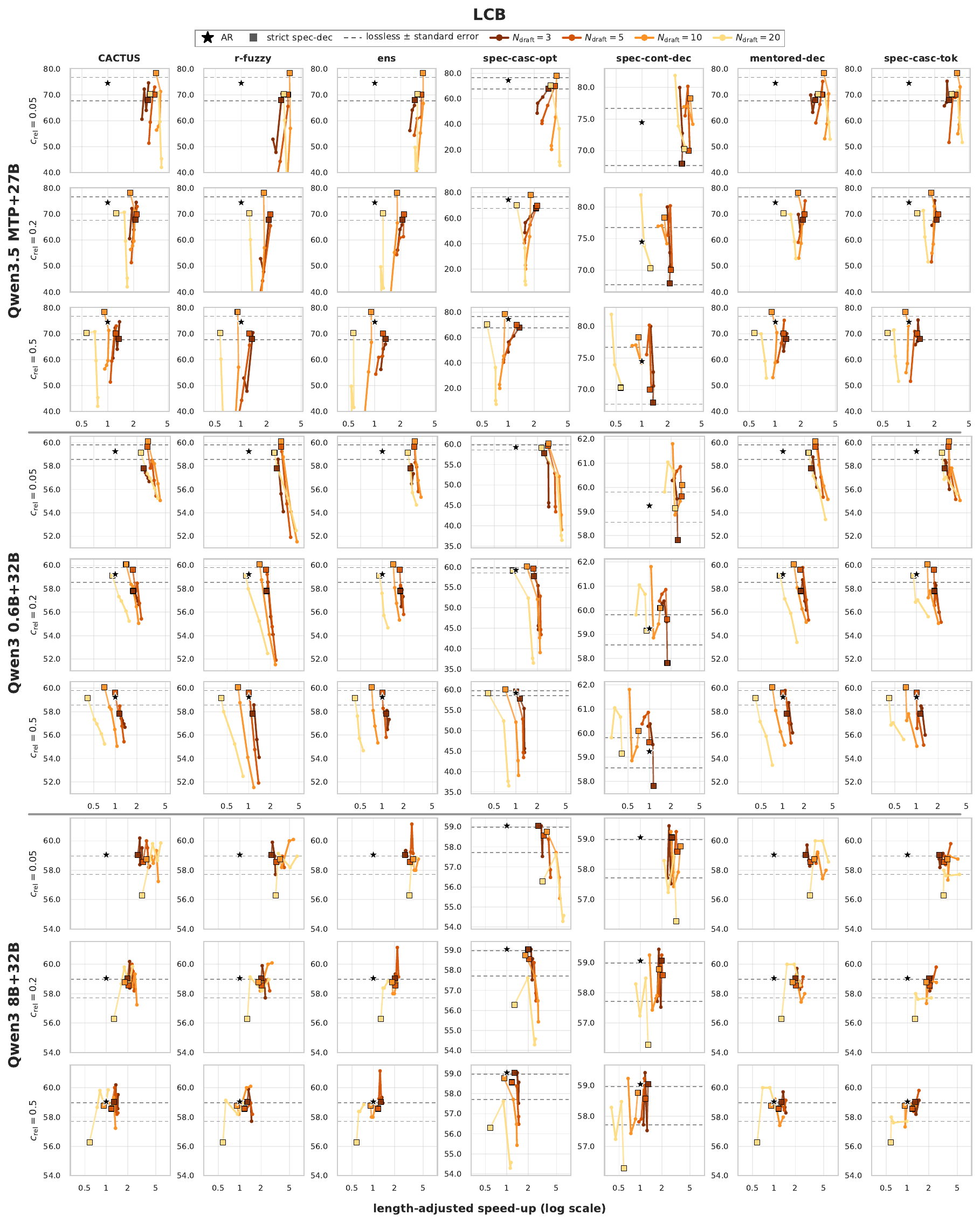}
    \caption{Capability-speed trade-offs on \texttt{LCB} for \texttt{relaxed} \texttt{spec-dec} methods across drafter+verifier pairs, relative drafter costs $c_\text{rel}$ and draft lengths $N_\text{draft}$. Each block shows one drafter+verifier pair, with method columns and $c_\text{rel}$ rows. The x-axis is length-adjusted speed-up (log scale) and the y-axis is task capability.}
    \label{fig:lcb-qwen-mtp-0p6b-8b-relaxed-supergrid}
\end{figure}

\begin{figure}[p]
    \centering
    \includegraphics[width=\linewidth]{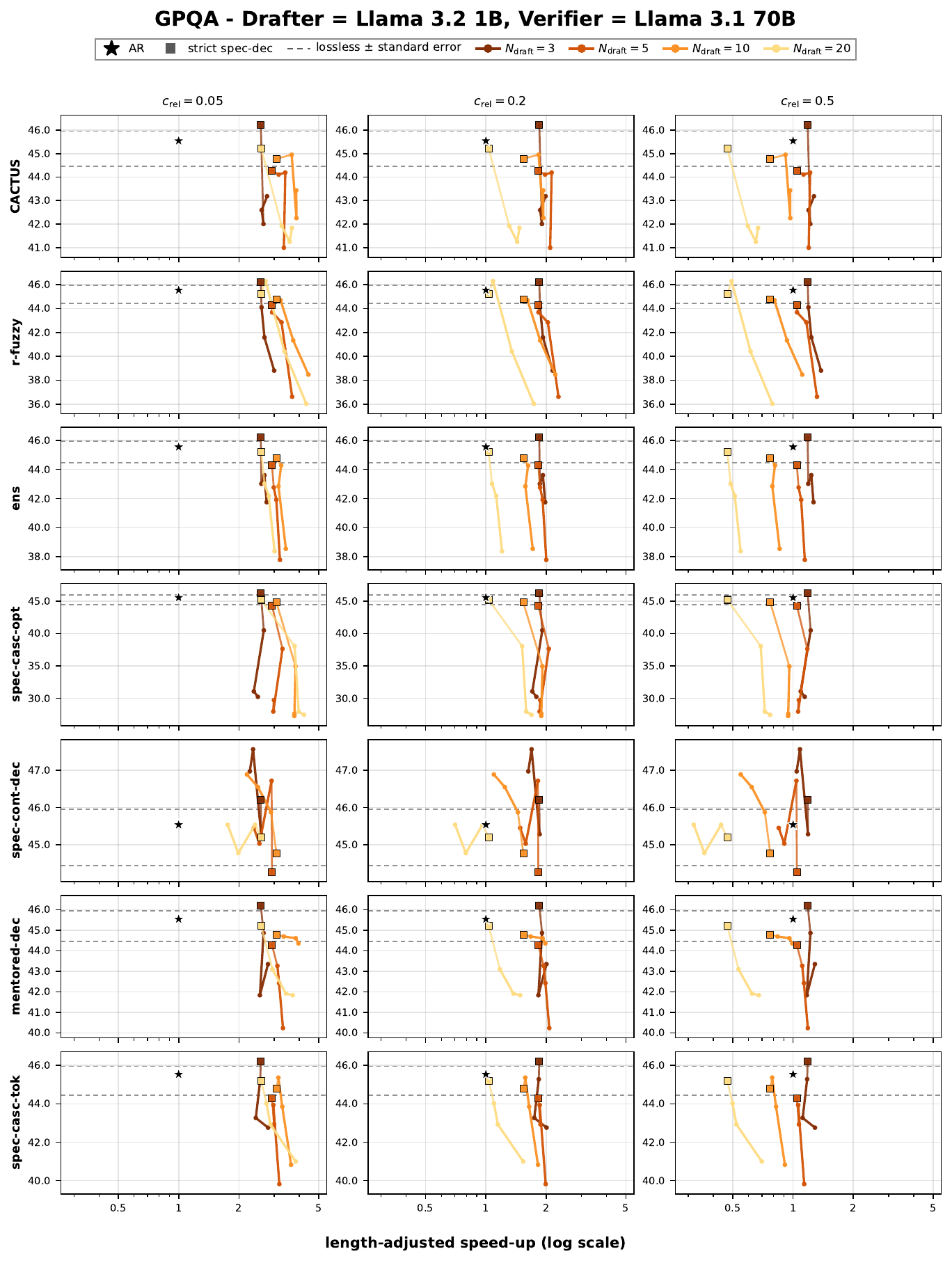}
    \caption{Capability-speed trade-offs on \texttt{GPQA} for \texttt{relaxed} \texttt{spec-dec} methods for Llama3, relative drafter costs $c_\text{rel}$ and draft lengths $N_\text{draft}$. The x-axis is length-adjusted speed-up (log scale) and the y-axis is task capability.}
    \label{fig:llama}
\end{figure}

Full empirical results are in \cref{fig:app-aime-qwen-mtp-0p6b-8b-relaxed-supergrid,fig:gpqa-qwen-mtp-0p6b-8b-relaxed-supergrid,fig:lcb-qwen-mtp-0p6b-8b-relaxed-supergrid,fig:llama}. These figures use the same capability-speed grid as the main body, without the highlighted annotations, so the broader behaviour across tasks can be inspected directly. Overall, broader results across \cref{fig:app-aime-qwen-mtp-0p6b-8b-relaxed-supergrid,fig:gpqa-qwen-mtp-0p6b-8b-relaxed-supergrid,fig:lcb-qwen-mtp-0p6b-8b-relaxed-supergrid,fig:llama} match the takeaways and findings in the main body of the paper.

\FloatBarrier

\section{Memory Cost Analysis}\label[appendix]{app:mem-cost}
\begin{figure}[t]
    \centering
    \includegraphics[width=\linewidth]{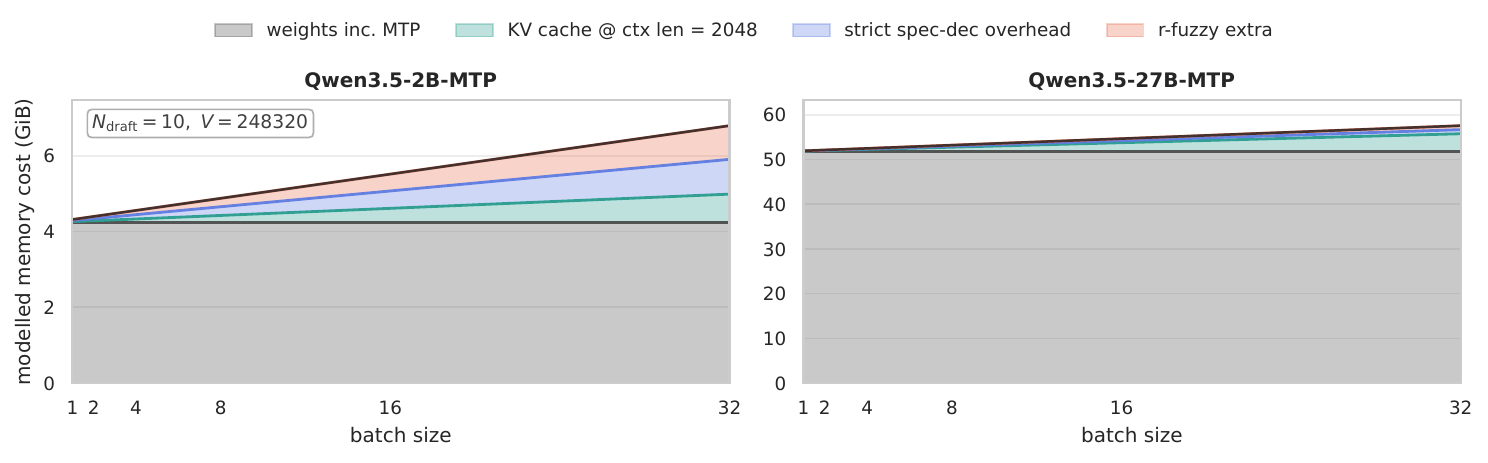}
    \caption{Modelled memory cost for native-MTP Qwen3.5 models as batch size increases. The decomposition separates checkpoint weights, KV cache, strict speculative-decoding overhead, and additional r-fuzzy overhead.}
    \label{fig:qwen35-memory-by-batch}
\end{figure}
\cref{fig:qwen35-memory-by-batch} shows how the memory cost of \texttt{r-fuzzy} for varying verifier size and batch size. \texttt{r-fuzzy} is the most memory hungry method in \cref{tab:relaxed-sd-taxonomy} as it requires additional \texttt{[batch\_size, vocab\_size, draft\_length]} tensors for (parallel) Jensen-Shannon Divergence calculation. We find that for the recent model Qwen3.5 we only see proportionally large memory overheads for a deployment setting of (small model, longer draft length, larger batch, shorter context).

\section{vLLM speculative decoding}\label[appendix]{app:vllm}
As shown in \cref{tab:vllm-drafter-cost-diagnostic}, on a single-GPU NVIDIA GH200 Grace Hopper Superchip with vLLM 0.20.1 and PyTorch 2.11.0+cu130, a small drafter is much cheaper when run as an ordinary autoregressive model than when used inside stock vLLM \texttt{spec-dec}. This gap is an implementation effect rather than a property of the drafter alone. In autoregressive decoding, vLLM can compile and capture the repeated decode loop, largely removing CPU-side scheduling and kernel-launch overheads. In separate-draft-model \texttt{spec-dec}, each round must instead coordinate drafting, verification, acceptance, resampling and cache updates, and this work is not collapsed into the same low-overhead loop. Our profiling indicates that the drafter path is therefore dominated by host-side coordination and many small GPU launches, making its effective cost much closer to the target-model token cost than its standalone autoregressive cost. We consequently treat stock vLLM wall-clock speed for separate autoregressive drafters as a diagnostic of current serving overhead, not as an inherent limit of \texttt{spec-dec}. This motivates our use of \cref{eq:speedup-approx,eq:gen-len-speedup} to explore potential speed-ups at different drafter costs.
\section{MTP Support}\label[appendix]{app:mtp-support}
We note that current production inference stacks do not always expose \texttt{spec-dec} with a separate autoregressive drafter once a model architecture has native MTP support. That is to say, frameworks sometimes \textit{force} users to use a bundled MTP drafter if it exists. In vLLM, models recognised as native-MTP-capable are redirected through an MTP-specific configuration path: a draft checkpoint whose architecture matches this family is rewritten to an MTP model class, and the speculative method is subsequently inferred as MTP rather than as a separate autoregressive draft model.\footnote{vLLM lists native-MTP model types in \texttt{MTPModelTypes}: \url{https://github.com/vllm-project/vllm/blob/b7f9b6ab271faa621f4cc438fd5ea7ecaf72db8e/vllm/config/speculative.py\#L34-L53}. Its speculative config override rewrites matching draft configs into MTP architectures, e.g. \url{https://github.com/vllm-project/vllm/blob/b7f9b6ab271faa621f4cc438fd5ea7ecaf72db8e/vllm/config/speculative.py\#L460-L468}, and later dispatches such draft configs to \texttt{method = "mtp"}: \url{https://github.com/vllm-project/vllm/blob/b7f9b6ab271faa621f4cc438fd5ea7ecaf72db8e/vllm/config/speculative.py\#L713-L716}.} SGLang implements a similar convention for its draft worker: the user-specified draft checkpoint is loaded in a mode that marks it as a drafter, and supported native-MTP architectures are then rewritten to the corresponding MTP implementation instead of retaining the full checkpoint as an independent autoregressive model.\footnote{SGLang's draft worker selects \texttt{speculative\_draft\_model\_path} while passing \texttt{is\_draft\_model=True}: \url{https://github.com/sgl-project/sglang/blob/50815d54a7b6502342aa037cf462cb1677190a82/python/sglang/srt/managers/tp_worker.py\#L329-L341}. Its model config then rewrites supported draft architectures to MTP variants, e.g. \url{https://github.com/sgl-project/sglang/blob/50815d54a7b6502342aa037cf462cb1677190a82/python/sglang/srt/configs/model_config.py\#L473-L488}, whose implementation consumes target-side speculative hidden states rather than operating as an ordinary independent LM: \url{https://github.com/sgl-project/sglang/blob/50815d54a7b6502342aa037cf462cb1677190a82/python/sglang/srt/models/qwen3_5_mtp.py\#L170-L181}.} Consequently, for such model families, supplying a smaller checkpoint through the standard draft-model interface does not provide a supported route to ordinary two-model \texttt{spec-dec}: the request is either diverted into the MTP implementation, or fails with load- or run-time errors when the resulting assumptions about target-side hidden states, weight names, dimensions, and cache structure are violated.
\end{document}